%% file: camera_ready.tex
\documentclass[sigconf]{aamas}

\usepackage{balance} 

\input{macros}

\setcopyright{ifaamas}
\acmConference[AAMAS '26]{Proc.\@ of the 25th International Conference
on Autonomous Agents and Multiagent Systems (AAMAS 2026)}{May 25 -- 29, 2026}
{Paphos, Cyprus}{C.~Amato, L.~Dennis, V.~Mascardi, J.~Thangarajah (eds.)}
\copyrightyear{2026}
\acmYear{2026}
\acmDOI{}
\acmPrice{}
\acmISBN{}

\acmSubmissionID{1456}

\ifthenelse{\boolean{arxivversion}}
{\title[Retrieval- and Argu\-mentation-Enhanced Multi-Agent LLMs for Judgmental Forecasting \\ (Extended Version with Supplementary Material)]{Retrieval- and Argu\-mentation-Enhanced Multi-Agent LLMs for 
Judgmental Forecasting \\ (Extended Version with Supplementary Material)}}
{\title[Retrieval- and Argu\-mentation-Enhanced Multi-Agent LLMs for Judgmental Forecasting]{Retrieval- and Argu\-mentation-Enhanced Multi-Agent LLMs for 
Judgmental Forecasting}}

\author{Deniz Gorur}
\affiliation{
  \institution{Imperial College London}
  \country{United Kingdom}}
\email{d.gorur22@imperial.ac.uk}

\author{Antonio Rago}
\affiliation{
  \institution{King's College London}
  \country{United Kingdom}}
\email{antonio.rago@kcl.ac.uk}

\author{Francesca Toni}
\affiliation{
  \institution{Imperial College London}
  \country{United Kingdom}}
\email{ft@imperial.ac.uk}

\begin{abstract}
Judgmental forecasting is the task of making 
predictions about future events based on human judgment.
This task can be seen as a form of claim verification, where the claim corresponds to a future event and the task is to assess the plausibility of that event.
In this paper,
we propose a novel 
multi-agent framework for claim verification, 
whereby different agents may disagree on claim veracity
and bring specific evidence for and against the claims,
represented as quantitative bipolar argumentation frameworks (QBAFs).
We then instantiate the framework for supporting claim verification,
with a variety of agents realised with Large Language Models (LLMs):
(1) ArgLLM agents, an existing approach for claim verification that generates and evaluates QBAFs
;
(2)  RbAM agents, whereby LLM-empowered Relation-based Argument Mining (RbAM) from external sources is used to generate QBAFs;
(3) RAG-ArgLLM agents, extending ArgLLM agents with a form of Retrieval-Augmented Generation (RAG) of arguments from external sources.
Finally, we conduct experiments with two standard  judgmental forecasting datasets,
with instances of our framework with two or three agents, empowered by six different base LLMs.
We observe that combining evidence from agents
can improve forecasting accuracy,
especially in the case of three agents,
while providing an explainable combination of evidence for claim verification.

\end{abstract}


\newcommand{\BibTeX}{\rm B\kern-.05em{\sc i\kern-.025em b}\kern-.08em\TeX}

\begin{document}


\pagestyle{fancy}
\fancyhead{}


\maketitle 


\section{Introduction}

\label{sec:introduction}
Judgmental forecasting is the task of making 
predictions about future events by reasoning over incomplete, uncertain, and often conflicting information \cite{lawrence2006judgmental,zellner2021human} based on the judgment of agents.
This 
task can be seen as a form of claim verification, where a future outcome is seen as a claim and the task is to assess the plausibility of that outcome.
The claim verification task involves determining whether a given claim is supported or refuted, 
often 
requiring the assessment of conflicting evidence. This evidence for verifying forecasting 
claims 
could be, for example, generated by LLMs \cite{Ge2025ResolvingEvidenceAutomatedFactChecking, Zheng2025Crave, Dmonte2024ClaimVerification} or obtained from other repositories. However, despite LLMs' remarkable 
capabilities across a range of tasks, they fall short 
here due to the fact that they may hallucinate or provide logically inconsistent outputs, and they cannot faithfully explain or allow for the contestation of their own outputs~\cite{Freedman2025ArgLLMs}.
For example, claims about future events may require up to date knowledge that a single LLM lacks, due to its incomplete training data, leading to unreliable or conflicting predictions \cite{Yue2024RAGFactVerification}.
One approach, which aims at targeting this issue, uses Argumentative LLMs (ArgLLMs) \cite{Freedman2025ArgLLMs}, which leverage techniques from computational argumentation (see \cite{AImagazine17,handbook} for overviews) to generate Quantitative Bipolar Argumentation Frameworks (QBAFs) that provide structured debates on the claim to be verified. 
In doing so, ArgLLMs output a transparent decision for the claims along with supporting and attacking arguments acting as evidence.
However, relying on a single ArgLLM is limiting, as its output is constrained by the knowledge and biases of its underlying LLM, potentially omitting crucial evidence.

\begin{figure*}
    \centering
    \ifthenelse{\boolean{arxivversion}}{
        \includegraphics[width=\linewidth]{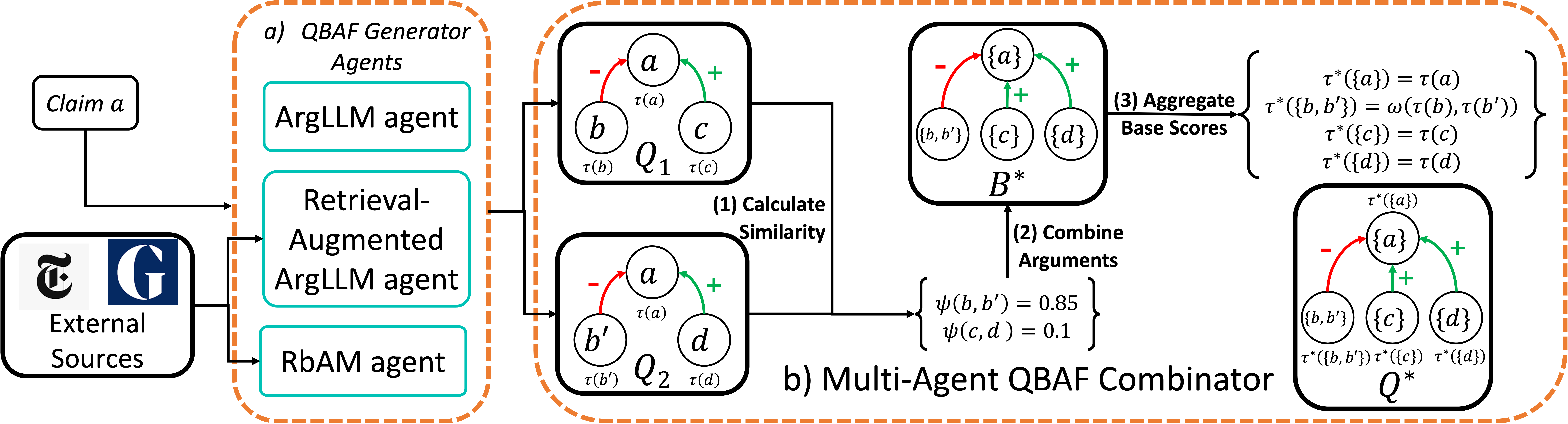}
    }{
        \includegraphics[width=0.95\linewidth]{images/pipeline.png}
    }
    \caption{The overall 
    pipeline.
    The `QBAF Generator Agents' module can be instantiated with ArgLLM agents (baseline) or our two RAG-based methods (§\ref{sec:llm_agents}): Retrieval-Augmented ArgLLM agents and RbAM agents, both taking in input external sources
    .
    The `Multi-Agent QBAF Combinator' (§\ref{sec:method}) module then takes the generated QBAFs (two in the figure, but our method applies to any number) 
    and (1) calculates similarity between arguments in the QBAFs, (2) combines similar arguments to obtain a single BAF $\mathcal{B}^*$, and (3) aggregates the 
    base scores of the combined arguments to obtain base scores $\tau^*$, leading to a combined QBAF $\mathcal{Q}^*$.
    }
    \label{fig:pipeline}
\end{figure*}

To overcome the limitations of a single-agent approach, we propose a novel multi-agent framework for claim verification that combines argumentative reasoning generated as QBAFs from multiple, independent agents into a single, more robust QBAF. As illustrated in Figure~\ref{fig:pipeline}, our \emph{Multi-Agent QBAF Combinator} module aggregates the outputs from several agents by measuring the semantic similarity \cite{Chandrasekaran2021SemanticSimilarity} between 
arguments, merging similar views to create a more robust framework. We chose QBAFs as they are core to ArgLLMs and have been deployed successfully in several applied settings, including judgmental forecasting~\cite{irwin2022forecasting} and decision-making~\cite{Evripidou2014Quaestio,Aurisicchio2015Arg&Dec}.


To instantiate our Multi-Agent QBAF Combinator module to support judgmental forecasting,
we consider two novel kinds of agents realised with LLMs, in addition to ArgLLM agents.
For these two novel kinds of agents, we leverage on 
the widely acknowledged fact that 
the integration of Retrieval-Augmented Generation
(RAG) \cite{Zhao2024RAGSurvey} enhances LLMs by incorporating external knowledge and solving issues like hallucination \cite{Yue2024RAGFactVerification,Ge2025ResolvingEvidenceAutomatedFactChecking}. We use
two novel types of RAG-based agents
: (1) using Relation-based Argument Mining (RbAM) \cite{Gorur2025LLMRbAM, Carstens2015TowardsRbAM} to identify supporting, attacking, or neither relations between 
retrieved evidence from sources and claims; and (2)  using the retrieved evidence from sources for generating supporting and attacking arguments, in the spirit of ArgLLMs.

Overall, our contributions, overviewed in Figure~\ref{fig:pipeline},
 are as follows:
\begin{enumerate}
    \item 
    \emph{Multi-Agent QBAF Combinator}: 
    a novel method to combine 
    independently generated QBAF outputs into a single 
    QBAF.
    \item 
    \emph{RbAM Agent}: 
    using relation-based argument mining to incorporate 
    evidence retrieved from external sources, directly as arguments.
    \item
    \emph{Retrieval-Augmented ArgLLM (RAG-ArgLLM) Agent}: 
    using evidence retrieved from external sources, in prompts.
    \item 
   Extensive experiments on two judgmental forecasting 
   datasets
   , demonstrating that our multi-agent framework, particularly with three distinct agents, can improve forecasting accuracy and provide a 
   transparent framework for claim verification in judgmental forecasting.
\end{enumerate}

\ifthenelse{\boolean{arxivversion}}{This pre-print is an extended version of \cite{aamasversion}, including supplementary material.}{An extended version of this paper \cite{arxivversion} includes additional information in the supplementary material.}


\section{Related Work}
\label{sec:related}


\paragraph{Claim verification with LLMs and RAG}
The field of claim verification has witnessed rapid advancements, particularly with LLMs, which can process extensive amounts of information and have remarkable generative capabilities.
Despite their strengths, LLMs encounter limitations in this task as they may generate conflicting evidence due to their diverse and incomplete training data \cite{Dmonte2024ClaimVerification}, which can lead to inaccurate predictions.
To address this, \citet{Zhang2023LLMFactVerificationNewsClaims} use a hierarchical step-by-step prompting method that breaks down the claim, which improves the accuracy of news claim verification.
Another solution is CRAVE \cite{Zheng2025Crave}, which improves accuracy by first eliminating ambiguities and retrieving evidence from external sources, and then reasoning through two conflicting perspectives.
Similarly, \citet{Ge2025ResolvingEvidenceAutomatedFactChecking} propose using RAG combined with LLMs to resolve conflicting evidence and improve claim verification accuracy.
Knowledge-based LLM approaches have been developed for structured claim verification, including generating argumentation frameworks to reason about claims (i.e. ArgLLMs \cite{Freedman2025ArgLLMs}), extracting argumentation frameworks from RAG sources to reason about claims (i.e. ArgRAG~\cite{Zhu2025ArgRAG}), generating first order logic representations to reason about claims \cite{Wang2023ExplainableClaimVerification}, and using multi-agent LLM systems for claim verification \cite{Fenza2025MultiLLMAgentsClaimVerification}.
We build on some of this earlier work on using LLMs to generate argumentation frameworks, but focus on combining the agents' generated frameworks. 

\paragraph{LLMs and Judgmental Forecasting}
Recent research has also explored the capabilities of LLMs in the context of judgmental forecasting, with mixed findings. Several studies have found that using LLMs directly, without external grounding, often fails to improve forecasting accuracy compared to human judgment \cite{Schoenegger2023large,Abolghasemi2023humans}. In contrast, \citet{Halawi2024approaching} demonstrated that LLMs augmented with RAG can approach the performance of expert human forecasters. Moreover, they show that combining forecasts from humans and LLMs achieves the best forecasting accuracy. Other work suggests that human forecasters advised by LLM-generated inputs can further enhance performance \cite{Schoenegger2024aiaugmented}.
Overall, while LLMs alone rarely outperform expert forecasters, RAG appears promising for achieving forecasting accuracy comparable to humans, which motivated us to incorporate it in our pipeline.


\paragraph{Combining Argumentation Frameworks}
In computational argumentation, there has been some work on merging argumentation frameworks.
\citet{Marquis2005MergingArgumentationSystems, Marquis2007MergingDungArgumentation} develop a merging pipeline that aligns agents' frameworks into a common domain, uses minimal edit distance operators to merge attack relations, and then votes on extensions to produce group level consensus arguments.
\citet{Cayrol2011WeightedArgumentationSystems} generalise this by introducing weighted argumentation frameworks that embed relative strengths of disagreement, producing a single weighted argumentation framework.
\citet{Leite2015MergingArgumentationSystems} merge argumentation frameworks using a semantic approach that selects arguments and attacks of an agent that vary the least from other agents.
\citet{Delobelle2016MergingAFs} merge the sets of acceptable arguments rather than merging argumentation frameworks directly.
Argumentative exchanges~\cite{Rago2023AX} allow combining knowledge from argumentation agents, but in a selected and distributed manner.
In this paper, we focus on merging QBAFs, a topic that, to our knowledge, has not been studied before.


\section{Preliminaries}
\label{sec:preliminaries}

A \emph{Bipolar Argumentation Framework (BAF)} \cite{Cayrol2005BAF} is a triple $\langle\mathcal{X}, \mathcal{A}, \mathcal{S} \rangle$, where: 
$\mathcal{X}$ is a set of \emph{arguments};
$\mathcal{A}\!\subseteq\!\mathcal{X}\times\mathcal{X}$ is a relation \emph{of attack}; and $\mathcal{S}\!\subseteq\!\mathcal{X}\times\mathcal{X}$ is a relation of \emph{support}, where $\mathcal{A}$ and $\mathcal{S}$ are disjoint ($\mathcal{A} \cap \mathcal{S} = \emptyset$).
A \emph{Quantitative Bipolar Argumentation Framework (QBAF)} \cite{Baroni2019gradual} is a tuple $\langle\mathcal{X}, \mathcal{A}, \mathcal{S}, \tau\rangle$ where: 
$\langle\mathcal{X}, \mathcal{A}, \mathcal{S} \rangle$  is a BAF and
$\tau\!:\!\mathcal{X}\!\rightarrow\![0,1]$ is a total function, with $\tau(a)$ the \emph{base score} of $a\!\in\!\mathcal{X}$. 
Arguments in QBAFs 
are then evaluated using \emph{gradual semantics}, i.e. total functions often in the form $\sigma\!:\!\mathcal{X}\!\rightarrow\![0,1]$, which assign a \emph{strength} to each argument. One such semantics is the \emph{Discontinuity-Free Quantitative Argumentation Debate (DF-QuAD)} \cite{Rago2016DFQuAD}, where for a given QBAF $\langle\mathcal{X},\mathcal{A},\mathcal{S},\tau\rangle$, for any $x \in \mathcal{X}$ with $n\geq 0$ attackers with strengths $v_1, \ldots, v_n$, $m\geq 0$ supporters with strengths $v_1', \ldots, v_m'$ and $\tau(x)=v_{0}$, DF-QuAD computes $x$'s strength as $\sigma(x)\!=\!\mathcal{C}(v_{0}, \mathcal{F}(v_{1}, \ldots, v_{n})$, $\mathcal{F}(v_{1}', \ldots, v_{m}'))$, where,
for any $w_{1}, \ldots, w_{k}$,  $\mathcal{F}(w_{1}, \ldots, w_{k})$ is $0$
if $k\!=\!0$  and $1 - \prod_{i=1}^k (|1-w_{i}|)$
otherwise, while
$\mathcal{C}$ is defined as follows: for $v_{a}=\mathcal{F}(v_{1}, \ldots, v_{n})$ and $v_{s}=\mathcal{F}(v_{1}', \ldots, v_{m}')$,
if $v_a=v_s$ then $\mathcal{C}(v_{0}, v_{a}, v_{s})=v_0$; else if $v_a > v_{s}$ then $\mathcal{C}(v_{0}, v_{a}, v_{s})=v_{0} - (v_{0}\cdot|v_{s}-v_{a}|)$; otherwise $\mathcal{C}(v_{0}, v_{a}, v_{s})=v_{0} + ((1-v_{0})\cdot|v_{s}-v_{a}|)$.

In this paper, we adapt the notation of a BAF/QBAF for an argument $a\in\mathcal{X}$ from
\citet{Rago2023AX}, as follows.
For $\mathcal{Q}\!=\!\langle \mathcal{X},\mathcal{A},\mathcal{S},\tau \rangle$ a QBAF and $\mathcal{B}\!=\!\langle \mathcal{X},\mathcal{A},\mathcal{S} \rangle$ a BAF, for any $\alpha, \beta \!\in \!\mathcal{X}$, let a \emph{path} from $\alpha$ to $\beta$ be 
$p\!=\!\langle(\alpha_0,\alpha_1), \ldots, (\alpha_{n-1}, \alpha_{n})\rangle$ for some $n\!>\!0$ (referred to as the \emph{length} of 
$p$, denoted $|p|$) where $\alpha_0 = \alpha$, $\alpha_n = \beta$ and, for any $1 \leq i \leq n$, $(\alpha_{i-1}, \alpha_{i}) \in \mathcal{A} \cup \mathcal{S}$. Let $\Path(\alpha,\beta)$ and $|\Path(\alpha,\beta)|$ indicate the set of all paths from $\alpha$ to $\beta$,
and the number of paths in $\Path(\alpha,\beta)$, respectively.
Then, for $\alpha^* \in \mathcal{X}$,
$\mathcal{Q}/\mathcal{B}$  is a \emph{BAF/QBAF for $\alpha^*$} iff 
    (i) $\forall_{\alpha \in \mathcal{X}} \Path(\alpha^*, \alpha)=\emptyset$;
    (ii) $\forall_{\alpha \!\in \!\mathcal{X} \setminus \{\alpha^*\}} |\Path(\alpha,\alpha^*)| = 1$;
    (iii) $\forall_{\alpha \in \mathcal{X}} \Path(\alpha, \alpha)=\emptyset$.
In essence, a BAF/QBAF for $\alpha^*$ is essentially a tree with $\alpha^*$ as the root, instead of a multi-tree considered in \cite{Rago2023AX}.

We also use \emph{pro/con} arguments in QBAFs as in \cite{Rago2023AX}.
\label{def:procon}
Let $\mathcal{Q}$ be a QBAF for $\alpha^*$. Then, the
\emph{pro arguments} and \emph{con arguments} for $\mathcal{Q}$ are, respectively:
$\Pro(\mathcal{Q}) \!\!=\!\! \{ \alpha \!\!\in\!\! \mathcal{X} | \exists p \!\in\! \Path(\alpha,\alpha^*), \text{where } | p \cap \mathcal{A} | \text{ is even} \}$;
$\Con(\mathcal{Q}) \!\!= \!\!\{ \alpha \!\!\in\!\! \mathcal{X} | \exists p \!\in\! \Path(\alpha,\alpha^*), \text{where } | p \cap \mathcal{A} | \text{ is odd} \}$.



\section{Multi-
Agent Claim Verification}
\label{sec:method}
To verify 
claims using multiple agents, we propose a method to combine the outputs of independently-generated QBAFs. Our method produces a combined QBAF by clustering the arguments across the  independently-generated QBAFs using a similarity measure, followed by aggregating their base scores. This allows us to capture a 
diverse argumentative perspective.
We assume that the independently-generated  QBAFs are such that there is no $(x,y)$ that is an attack in one of the  QBAFs and a support in another, i.e. assuming a lingua franca for the relations as in \cite{Rago2023AX}
.

To determine whether arguments should be combined, we first need a formal notion of their similarity. We thus define a similarity function for clustering arguments.

\begin{definition}
\label{def:similarity}
Let $\mathcal{X}$ be a set of arguments. 
A \emph{similarity function}
$\Psi: \mathcal{X}\times \mathcal{X} \rightarrow [0,1]$
is such that, for $x \in \mathcal{X}$, $\Psi(x,x)=1$
and for $x,y \in \mathcal{X}$,
$\Psi(x,y)=\Psi(y,x)$.
\end{definition}

This definition ensures that the similarity of an argument to itself is maximal and that the similarity function is independent of the order in which arguments are compared.

To aggregate multiple base scores assigned to similar arguments, we define a base score aggregation function. This function allows for the aggregation of vectors of base scores into a single representative value.

\begin{definition}
\label{def:aggregation}
Let $K \in \mathbb{N}$. A \emph{base score aggregation function w.r.t. $K$}, $\omega:{\bigcup_{k=1}^K} [0,1]^k \rightarrow [0,1]$
is such that:
\begin{enumerate}
    \item for any $v \in [0,1]^k$, for any permutation $v'$ of the elements of $v$, $\omega(v')=\omega(v)$ (\emph{order-independence});
    \item for any $v
    \in [0,1]^k$, $\min(v) \leq \omega(v) \leq \max(v)$  (\emph{boundedness});
    \item for any $v_i \in [0,1]$, $\omega((v_i, \ldots, v_i)) = v_i$ (\emph{idempotence});
    \item for any $v,v' \!\in\! [0,1]^k$ with $v\!=\!(v_1, \ldots, v_k)$
    and $v'\!=\!(v_1', \ldots, v_k')$, if $\forall_{i \in \{1, \ldots, k\}}(v_i \leq v_i')$ then $\omega(v) \leq \omega(v')$ (\emph{monotonicity}).
\end{enumerate}
\end{definition}

This definition ensures that base score aggregation functions 
do not depend on the order in which the elements in their input are presented, always return a number which lies within the range of the provided base scores, if all elements in the input are the same then the output is the common element, and it is monotonic w.r.t. the inputs.
We consider two instantiations of the base score aggregation function $\omega$:
    average aggregation, i.e. $\omega_{avg}((v_1, \ldots, v_k))=\frac{1}{k}\sum_{i=1}^k v_i$; and
    maximum aggregation, i.e. $\omega_{max}((v_1, \ldots, v_k))=\max_{i=1}^k v_i$.
It is easy to see that both aggregation functions satisfy Definition~\ref{def:aggregation} (in particular both functions are order-independent, bounded, idempotent, and monotonic --
\ifthenelse{\boolean{arxivversion}}{see Appendix~\ref{app:proof} for the proof)}{see \cite{arxivversion})}.

Next, we define what it means to aggregate a set of QBAFs into a combined QBAF, given a similarity function and a base score aggregation function. This combined QBAF captures argument clusters, relations between them, and their aggregated base scores.

\begin{definition}
\label{def:combine}
Let  $\mathcal{Q}_1, \dots, \mathcal{Q}_n$  be $n > 1$ QBAFs, where, for $i \in \{1,\dots,n\}$, $\mathcal{Q}_i=\langle\mathcal{X}_i, \mathcal{A}_i, \mathcal{S}_i, \tau_i\rangle$.
Let $\mathcal{X}=\bigcup_{i=1}^n \mathcal{X}_i$, $\mathcal{A}=\bigcup_{i=1}^n \mathcal{A}_i$, and $\mathcal{S}=\bigcup_{i=1}^n \mathcal{S}_i$.
Let $\Psi: \mathcal{X}\times \mathcal{X} \rightarrow [0,1]$ be a similarity function, and  
$\delta\in[0,1]$ be a similarity threshold.
Let 
$\omega:{\bigcup_{k=1}^K} [0,1]^k \rightarrow [0,1]$
be a base score aggregation function wrt $K=|\mathcal{X}|$. 
Then, 
the \emph{combined QBAF} $\mathcal{Q}^*=\langle\mathcal{X}^*, \mathcal{A}^*, \mathcal{S}^*, \tau^*\rangle$ is 
as follows:

\begin{itemize}
\item $\mathcal{X}^*\subseteq 2^{\mathcal{X}}$ satisfies the following properties:

\begin{enumerate}
\item $\forall x\in \mathcal{X}$, $\exists_1 x^*\in\mathcal{X}^*$ such that $x\in x^*$

\item $\forall x,y\in \mathcal{X}$,
$\exists x^*\in\mathcal{X}^*$ such that $x,y\in x^*$ iff (i) $\exists z^*\in \mathcal{X}^*$ and $\exists z,z' \in z^*$ with $(x,z), (y,z') \in \mathcal{A}$ or $(x,z), (y,z') \in \mathcal{S}$, and (ii) $\Psi(x,y)\geq\delta$.
\end{enumerate}

\item $\mathcal{A}^*=\{(x^*,y^*)|\exists_{x \in x^*}\exists_{y \in y^*} \ \text{such that} \ (x,y) \in \mathcal{A}\}$;

\item $\mathcal{S}^*=\{(x^*,y^*)|\exists_{x \in x^*}\exists_{y \in y^*} \ \text{such that} \ (x,y) \in \mathcal{S}\}$;

\item $\forall x^*=\{x_1, \ldots, x_k\} \in \mathcal{X}^*$, $\tau^*(x^*)=\omega(\tau(x_1), \ldots, \tau(x_k))$.
\end{itemize}
\end{definition}

The first property ensures assignment of each argument $x\in\mathcal{X}$ to a unique cluster, forming a singleton if no other properties apply.
The second property ensures that any two arguments are in the same cluster if and only if they both have the same relation (attack or support) towards arguments in the same parent 
cluster and 
meet the threshold of the similarity function.
Combined QBAFs preserve relations by
adding relations between clusters
and aggregating the base scores 
according to the clusters.
Note that, not only each attack/support between clusters results from lifting an attack/support between arguments, but also each attack/support
between arguments is captured by some attack/support between clusters. Formally:

\begin{restatable}{lemma}{lifting}
\label{lemma:lifting}
    Let $x^*,y^* \in \mathcal{X}^*$. Then, $(x^*,y^*) \in \mathcal{A}^*$ (or $(x^*, y^*) \in \mathcal{S}^*$) iff $\exists x \in x^*$ and $\exists y \in y^*$ such that $\exists {i \in \{1,\dots,n\}}$ where $(x, y) \in \mathcal{A}_i$ (or $(x, y) \in \mathcal{S}_i$, respectively).
\end{restatable}

\begin{example}
    \label{ex:simple}
    Given two QBAFs, 
    $\mathcal{Q}_1\!=\!\langle\mathcal{X}_1\!=\!\{a,b,c\}, \mathcal{A}_1\!=\!\{(b,a)\}, \\ \mathcal{S}_1\!=\!\{(c,a)\}, \tau_1(a)\!=\!0.5, \tau_1(b)\!=\!0.2, \tau_1(c)\!=\!0.8 \rangle$ and $\mathcal{Q}_2\!=\!\langle\mathcal{X}_1\!=\!\{a,b',d,e,e'\}, \mathcal{A}_1\!=\!\{(b',a)\}, \mathcal{S}_1\!=\!\{(d,a),(e,b'), (e',b')\}, \tau_2(a)\!=\!0.5, \tau_2(b')\!=\!0.7, \tau_2(d)\!=\!0.4, \tau_2(e)\!=\!0.3, \tau(e')\!=\!0.1 \rangle$.
    We assume a similarity function $\Psi$ such that $\Psi(b,b') \!=\! 0.9$, $\Psi(e,e') \!=\! 0.6$, and all other possible argument pairs have similarity below the threshold $\delta \!=\! 0.5$.
    Let $\omega$ be the arithmetic mean $\omega_{avg}$.
    We construct the combined QBAF $\mathcal{Q}^*\!=\!\langle\mathcal{X}^*, \mathcal{A}^*, \mathcal{S}^*, \tau^*\rangle$ as follows:
    $\mathcal{X}^*=\{
        x_1^* \!=\! \{a\} \ \text{(singleton, no parent argument)}; \
        x_2^* \!=\! \{b, b'\} \ (\text{since} \ \Psi(b, b') \!=\! 0.9 > \delta); \
        x_3^* \!=\! \{c\} \ \text{(singleton, no similar arguments)}; \
        x_4^* \!=\! \{d\} \text{(single-} \\ \text{ton, no similar arguments)}; \
        x_5^* \!=\! \{e, e'\} \ (\text{since} \ \Psi(e, e') \!=\! 0.6 > \delta)
        \}$.
        $\mathcal{A}^*\!=\!\{
        (x_2^*, x_1^*) \ \text{since} \ (b,a) \in \mathcal{A}_1, (b',a) \in \mathcal{A}_2; \
        (x_5^*, x_2^*) \text{since} \ (e,b') \in \mathcal{A}_2, (e',b') \in \mathcal{A}_2
        \}$.
        $\mathcal{S}^*\!=\!\{
        (x_3^*, x_1^*) \ \text{since} \ (c,a) \in \mathcal{S}_1; \
        (x_4^*, x_1^*) \ \text{since} \\ (d,a) \in \mathcal{S}_2
        \}$.
        $\tau^*(x_1^*) \!=\! \omega(0.5, 0.5) \!=\! 0.5; \
        \tau^*(x_2^*) \!=\! \omega(0.2, 0.7) \!=\! 0.45; \
        \tau^*(x_3^*) \!=\! 0.8; \
        \tau^*(x_4^*) \!=\! 0.4; \
        \tau^*(x_5^*) \!=\! \omega(0.3, 0.1) \!=\! 0.2$.
\end{example}


\begin{figure}[htp!]
    \centering
    \ifthenelse{\boolean{arxivversion}}{
    \includegraphics[width=\linewidth]{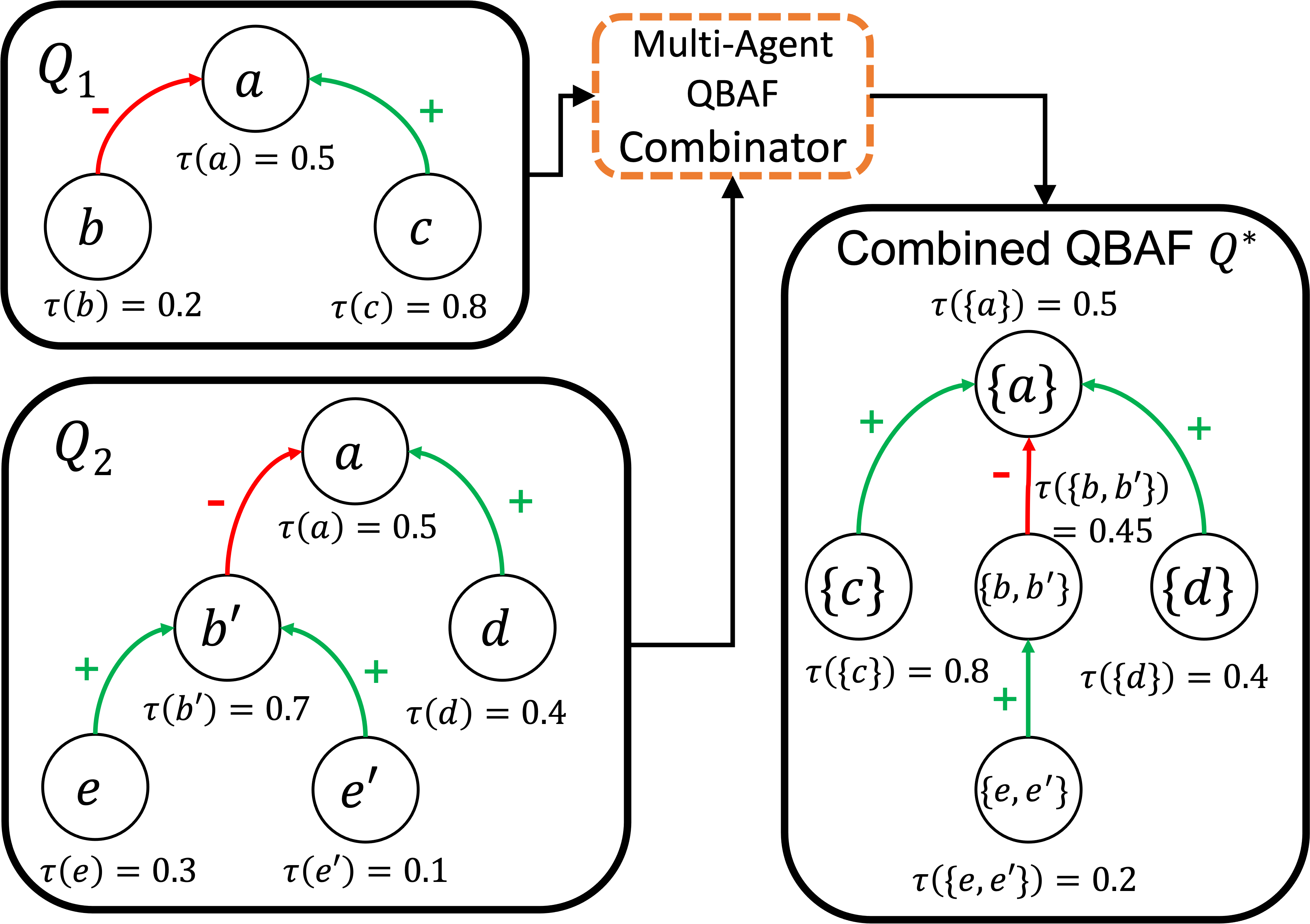}
}{
    \includegraphics[width=0.95\linewidth]{images/simple.png}
}
    \caption{The Multi-Agent QBAF Combinator takes two initial QBAFs, $\mathcal{Q}_1$ (top-left) and $\mathcal{Q}_2$ (bottom-left), as input and outputs a single, merged QBAF $\mathcal{Q}^*$ (right).}
    \label{fig:simple}
\end{figure}

\begin{restatable}{proposition}{qbaf}
\label{prop:qbaf}
    Let $\mathcal{Q}_1, \ldots, \mathcal{Q}_n$ be QBAFs, for $n > 1$, where, for $i \in \{1, \ldots, n\}$, $\mathcal{Q}_i=\langle\mathcal{X}_i, \mathcal{A}_i, \mathcal{S}_i, \tau_i\rangle$. Then, the combined QBAF $\mathcal{Q}^*=\langle\mathcal{X}^*, \mathcal{A}^*, \mathcal{S}^*, \tau^*\rangle$ 
    is a QBAF. 
\end{restatable}

\begin{restatable}{proposition}{qbafforf}
\label{prop:qbaf_for_f}
    Let $\mathcal{Q}_1, \ldots, \mathcal{Q}_n$ be QBAFs for the same argument $f$
    , for $n > 1$, where, for $i \in \{1, \ldots, n\}$,  $\mathcal{Q}_i=\langle\mathcal{X}_i, \mathcal{A}_i, \mathcal{S}_i, \tau_i\rangle$. Then, the combined QBAF $\mathcal{Q}^*=\langle\mathcal{X}^*, \mathcal{A}^*, \mathcal{S}^*, \tau^*\rangle$ 
    is a QBAF for $\{f\}$. 
\end{restatable}

\begin{restatable}{proposition}{qbafprocon}
\label{prop:qbaf_pro_con}
    Let $\mathcal{Q}_1, \ldots, \mathcal{Q}_n$ be QBAFs for the same argument $f$, for $n > 1$, where, for $i \in \{1, \ldots, n\}$,  $\mathcal{Q}_i=\langle\mathcal{X}_i, \mathcal{A}_i, \mathcal{S}_i, \tau_i\rangle$.
    Then, the combined QBAF $\mathcal{Q}^*=\langle\mathcal{X}^*, \mathcal{A}^*, \mathcal{S}^*, \tau^*\rangle$ from Definition~\ref{def:combine} preserves pro/con arguments in the original QBAFs, i.e., for  
    $\Pro(Q_i)$ ($\Con(Q_i)$) the set of pro-arguments (con-arguments, respectively) for $\mathcal{Q}^*$:
    
    \hspace*{-0.4cm}
    \(\forall_{x^* \in \mathcal{X}^*} \left( x^* \in \Pro(\mathcal{Q}^*) \Leftrightarrow (\exists_{x \in x^*, i \in \{1, \ldots, n\}} x \in \Pro(\mathcal{Q}_i)) \right)\), and

    \hspace*{-0.4cm}
    \(\forall_{x^* \in \mathcal{X}^*} \left( x^* \in \Con(\mathcal{Q}^*) \Leftrightarrow (\exists_{x \in x^*, i \in \{1, \ldots, n\}} x \in \Con(\mathcal{Q}_i)) \right)\).


    


\end{restatable}

Note that arguments without any similarities with other arguments (e.g. $d$ in Example~\ref{ex:simple}) 
form singleton clusters (e.g.  $\{e\}$ in the same example). 
Note that we do not cluster arguments with different parents or with the same parent but in different relations. Thus, in Example~\ref{ex:simple}, we do not cluster $c$ and $e$ (as they do not share a parent) or $b$ and $d$ (as they have opposite stances for the same parent) even if they are similar.
Clustered arguments are made to share the same incoming relations in the combined QBAF. Thus, in Example~\ref{ex:simple}, $b$ and $b'$ in the cluster $\{b,b'\}$ share supporter $\{e,e'\}$.

In the remainder of the paper, we assume that the QBAFs are trees as in Example~\ref{ex:simple} (and thus acyclic graphs) for claim $a$.

To construct a combined QBAF satisfying
Definition~\ref{def:combine}, we define a bottom-up algorithm that clusters arguments layer by layer, based on the similarity function. The Multi-Agent QBAF Combinator algorithm starts from a given claim argument $a$ and then clusters its supporters and attackers, recursively clustering their children, until the maximum depth of the input QBAFs is reached. For each cluster, the algorithm aggregates their base scores.

\begin{algorithm}[htp!]
\caption{Multi-Agent QBAF Combinator}
\label{alg:cap}
\begin{algorithmic}[1]
\REQUIRE $F = \{\mathcal{Q}_1, \dots, \mathcal{Q}_n\}$: QBAFs for claim $a$, $n>1$ 
\REQUIRE $\Psi$: similarity function
\REQUIRE $\delta$: similarity threshold
\REQUIRE $\omega$: base score aggregation function

\STATE Initialize empty QBAF $\mathcal{Q}^* \gets \langle \{\{x\} \mid x \in \mathcal{X}\} 
,\emptyset, \emptyset, \tau^*(a)=\omega(\tau_1(a), \ldots, \tau_n(a)) \rangle$
\STATE previous layer $\gets \{a\}$
\FOR{$d \gets 1$ to max depth level($\{\mathcal{Q}_1, \dots, \mathcal{Q}_n\}$)}
    \STATE merged $\gets \{\}$
    \FOR{$z^* \gets$ previous layer}
        \FORALL{pairs $(x, y)$ where $x,y\in\mathcal{X}$ such that $(x, z), (y, z') \in \mathcal{A}$ or $(x, z), (y, z') \in \mathcal{S}$ where $z,z' \in z^*$}
            \IF{$\Psi(x, y) \geq \delta$}
                \STATE 
                merge $x^*$ and $y^*$, where $x \in x^*$ and $y \in y^*$
                \STATE add $x$, $y$ to same cluster in merged
            \ELSE
                \STATE add $x$ to cluster in merged
                \STATE add $y$ to another cluster in merged
            \ENDIF
        \ENDFOR
    \ENDFOR
    \FOR{$z^* \gets$ previous layer and $x^*\gets$ merged}
        \STATE set $\tau^*(x^*)=\omega(\tau(x)\mid x\in x^*)$
        \IF{$\exists_{x\in x^*}(x,z) \in \mathcal{A}$ such that $z\in z^*$}
            \STATE add $(x^*, z^*)$ to $\mathcal{A}^*$ 
        \ENDIF
        \IF{$\exists_{x\in x^*}(x,z) \in \mathcal{S}$ such that $z\in z^*$}
            \STATE add $(x^*, z^*)$ to $\mathcal{S}^*$
        \ENDIF
    \ENDFOR
    
    \STATE previous layer $\gets 
    $ merged
\ENDFOR

\RETURN $\mathcal{Q}^*$: combined QBAF
\end{algorithmic}
\end{algorithm}

Lines 1-2 initialise the combined QBAF $\mathcal{Q}^*$ with singleton clusters for every argument in the union QBAF $x\in\mathcal{X}$. We set the `previous layer' to $\{a\}$ the root, which represents the parent arguments in the previous layer.
The algorithm then executes recursively layer by layer, Lines 4-12 merge clusters in $\mathcal{X}^*$ and build a set of clusters `merged' at the current depth. Arguments are in the same cluster if they support or attack the same parent argument and are similar. Otherwise, they are placed in separate clusters. Lines 13-18 add the relation between `merged' clusters to the combined QBAF, and assign clusters their aggregated base score. Line 19 updates `previous layer' to contain all merged arguments in the previous layer, enabling the algorithm to iterate to the next layer.

\begin{example}
We apply Algorithm~\ref{alg:cap} to Example~\ref{ex:simple}'s $\mathcal{Q}_1$ and $\mathcal{Q}_2$.

In line 1, the combined QBAF $\mathcal{Q}^*$ is initialized with singleton clusters of all arguments in the union QBAF: $\{\{a\},\{b\},\{c\},\{b'\},\{d\},\{e\}, \\ \{e'\}\}=\mathcal{X}^*$, and the base score of $a$ is set to $\tau^*(a)\!=\!\omega(\tau_1(a), \ldots, \tau_n(a))$. In line 2, the `previous layer' is set to $\{a\}$.

At depth 1, `merged' is set to $\{\}$ in line 4. Lines 5-12 process
all arguments in `previous layer' ($a$ in depth 1). Lines 6-12 goes through $(b,b'),(c,d)$ as $(b,a),(b',a)\in\mathcal{A}$ and $(c,a),(d,a)\in\mathcal{S}$. $\Psi(b, b') > 0.5$ so `merged'=$\{\{b,b'\}\}$. $\Psi(c, d) < 0.5$ so `merged'$\!=\! \{\{b,b'\}, \{c\}, \{d\}\}$. Line 8 merges clusters in the combined QBAF, $\{\{b,b'\}, \{c\}, \{d\}\}\subset\mathcal{X}^*$.

Lines 13-18 goes through all merged clusters $\{b,b'\}, \{c\}, \{d\}$ and all arguments in `previous layer'. Then, the base scores are added in line 14, $\tau^*(\{b,b'\})=\omega(\tau(b), \tau(b'))=\omega(0.5,0.5)=0.5$, $\tau^*(\{c\})=0.8$, $\tau^*(\{d\})=0.4$. Lines 15-18 adds support (attack) relations if there is an argument in the cluster that is supporting (attacking) an argument in the `previous layer'. So, $\{(\{b,b'\}, a)\}\subset\mathcal{A}^*$ as $(b,a)\in\mathcal{A}$, and $\{(\{c\}, a), (\{d\}, a)\}\subset\mathcal{S}^*$ as $(c,a),(d,a)\in\mathcal{S}$.

Before starting depth 2, line 19 assigns the union of all arguments that were in `merged' to `previous layer'. `Previous layer'$=\{b,b',c,d\}$.

`merged' is set to $\{\}$ in line 4. Lines 5-12 goes through all arguments in `previous layer' ($b,b',c,d$ in depth 1). Lines 6-12 goes through $(e,e')$ as $(e,b'),(e',b')\in\mathcal{S}$. $\Psi(e, e') > 0.5$ so $\text{`merged'}=\{\{e,e'\}\}$. Line 8 merges the clusters in the combined QBAF, $\{\{e,e'\}\}\subset\mathcal{X}^*$.

Lines 13-18 goes through all merged clusters $\{e,e'\}$ and all arguments in `previous layer'. Then, the base scores are added in line 14, $\tau^*(\{e,e'\})=\omega(\tau(e), \tau(e))=\omega(0.3,0.1)=0.2$. Lines 15-18 $\{(\{e,e'\}, \{b,b'\})\}\subset\mathcal{A}^*$ as $(e,b)\in\mathcal{S}$.

Finally, the algorithm returns $\mathcal{Q}^*$ as the combined QBAF.
\end{example}

\begin{restatable}{proposition}{algorithm}
Algorithm~\ref{alg:cap} terminates and returns a combined QBAF in polynomial time. 
\end{restatable}

Note that we have defined the combination process algorithmically, however the Multi-Agent QBAF Combinator can be seen as a {\em judge agent}. In this view, the generating agents act as independent experts providing potentially conflicting evidence ($\mathcal{Q}_1, \ldots, \mathcal{Q}_n$) and the 
judge reconciles these opinions
.


\section{LLM-Agents for QBAF Generation}
\label{sec:llm_agents}
In this section, we describe the three LLM-agent variants we consider: Argumentative LLM (ArgLLM) agents, Retrieval-Augmented ArgLLM (RAG-ArgLLM) agents (extending ArgLLMs with RAG, to incorporate external textual evidence),
and RbAM agents (complementary approach, which constructs a QBAF by directly using the external sources as arguments and classifying these arguments stances towards the claim). Each of these agents independently generates QBAFs.
We use approaches based on ArgLLMs in this paper instead of ArgRAG because the QBAFs generated for our multi-agent
framework to combine QBAFs must be acyclic trees, while it is possible that ArgRAG could generate a QBAF with cycles.


\subsection{ArgLLM Agent}
The ArgLLM \cite{Freedman2025ArgLLMs} agent extracts QBAFs from LLMs and formally reasons over them using gradual semantics. The ArgLLMs agent
has three components: 
(1) {\em Argument Generation}, produces
a BAF $\mathcal{B}$ for a given claim $a$, 
given a generative model $G$, and parameters for argument generation $\theta$, such as the prompt used, and the number of arguments to be generated in depth and breadth, denoted formally as $\Gamma(x,G,\theta)\!\rightarrow\! \mathcal{B}$;
(2) {\em Intrinsic Argument Strength Attribution}, 
assigns base scores to the BAF $\mathcal{B}$ using an evaluative function $E$ to obtain a QBAF $\mathcal{Q}$, defined as
$\mathcal{E}(\mathcal{B},E)\!\rightarrow\! \mathcal{Q}$;
(3) {Argument Strength Calculation}, 
applies a gradual semantics $\sigma$ to the QBAF $\mathcal{Q}$ to obtain an assessment for the claim $a$, denoted as
$\Sigma(a, \mathcal{Q}, \sigma)\!\rightarrow\!\sigma(a)$.

\subsection{RAG-ArgLLM Agent}
To strengthen the factual grounding of generated arguments, we extend the ArgLLM agent with RAG. In this work, ArgLLM agents are prompted with textual evidence from external sources, allowing it to generate more complete and informed arguments.

Let $T=\{t_1,...,t_k\}$ be a set of textual evidence relevant to a claim $a$, retrieved from a collection of sources, and let $G$ be a generative model.
We define \emph{Retrieval-Augmented ArgLLMs} as one in which $\theta$ in the argument generation function is a modified prompt, \emph{RAG prompt for ArgLLMs}, which incorporates the retrieved evidence $T$.

\begin{figure}[htp!]
    \centering
    \ifthenelse{\boolean{arxivversion}}{
        \includegraphics[width=\linewidth]{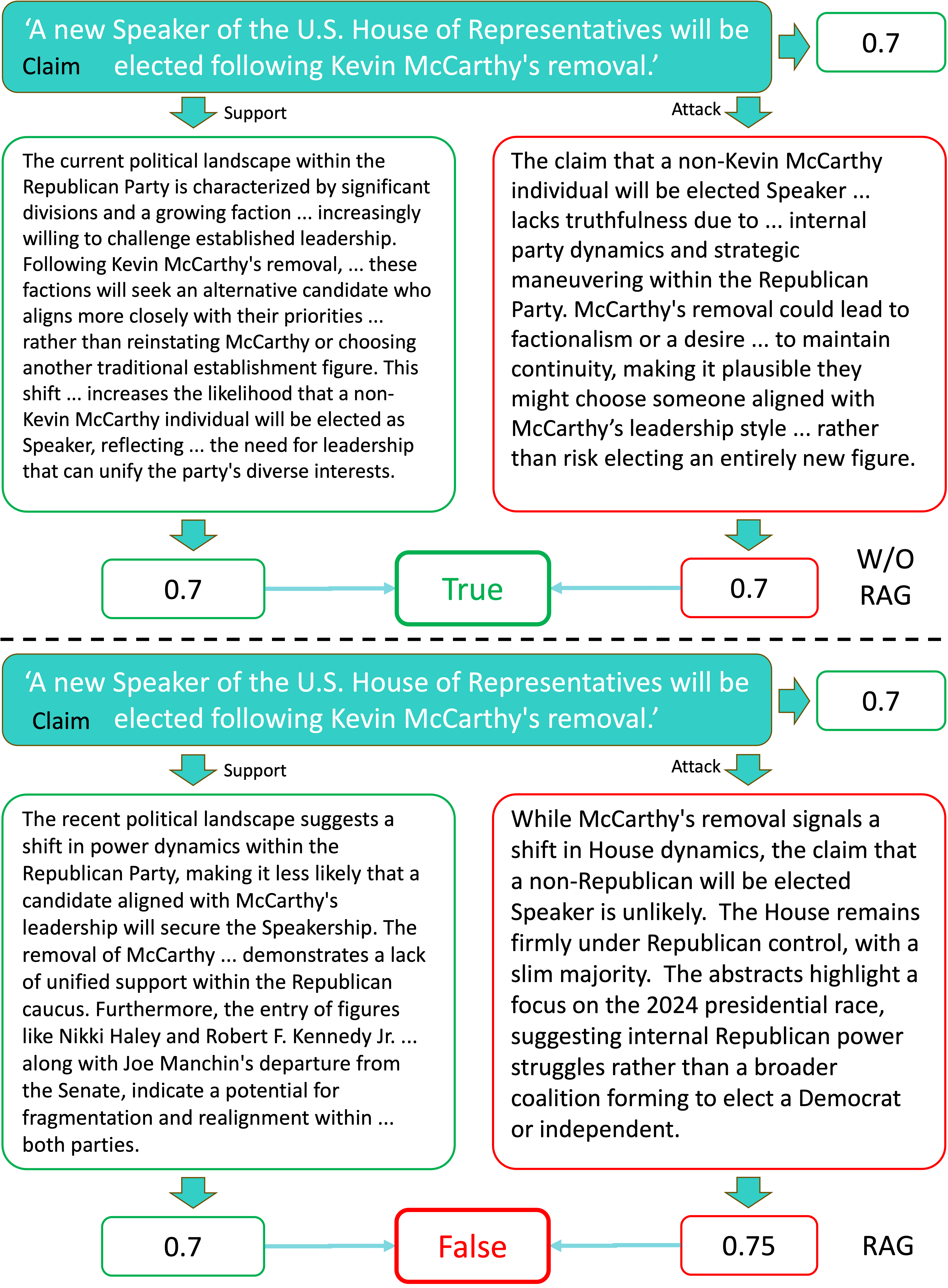}
    }{
        \includegraphics[width=0.95\linewidth]{images/rag.png}
    }
    \caption{An example of how an RAG-ArgLLM agent (bottom) can improve the results compared to an ArgLLM agent (top). The example is taken from the Metaculus dataset and its claim is false. The ArgLLM agent incorrectly predicts it as false whereas the RAG-ArgLLM agent correctly predicts it.}
    \label{fig:rag}
\end{figure}

Figure~\ref{fig:rag} shows how RAG improves a result for a single claim.


\subsection{RbAM Agent}
In addition to RAG-ArgLLM agents, we introduce a complementary approach based on RbAM. In this setting, textual evidence is not used to generate new arguments but is instead seen as arguments. The RbAM component classifies the relation between each textual evidence and the claim as support, attack, or neither, yielding a QBAF in which the retrieved evidence supports or attacks the claim.


The \emph{RbAM Agent}, treats each retrieved evidence 
as an argument, and the RbAM component determines whether it is related to the claim $f$.
We adopt the best performing RbAM method from \cite{Gorur2025LLMRbAM}: a few-shot prompt-based classification approach using the Mixtral-8x7B \cite{jiang2024mixtral} model, which \citet{Gorur2025LLMRbAM} show significantly outperforms prior baselines on the RbAM task.


\section{Experimental Set-up}
\label{sec:experimental}
First, we evaluate our individual LLM-Agents introduced in §\ref{sec:llm_agents}:
(1) the ArgLLM agent (as a baseline),
(2) the RAG-ArgLLM agent, and
(3) the RbAM agent.

Then, we evaluate our multi-agent framework by combining the outputs of these agents in various configurations:
(1) pairs of ArgLLM agents across six base models,
(2) pairs of RAG-ArgLLM agents across six base models,
(3) two-agent combinations: 
combining RAG-ArgLLM agents that use different external sources (The NYTimes and The Guardian),
(4) a three-agent combination: combining the ArgLLM agent with the two different RAG-ArgLLM agents, each using a different external source.


In our experiments, we use the Jina-V3 embeddings \cite{Sturua2025JinaV3} with cosine similarity for the similarity function $\Psi$. This embedding model was selected due to its strong performance on semantic similarity tasks and compact size\footnote{At the time of experimentation, Jina-V3 ranked fourth on the MTEB leaderboard: \url{https://huggingface.co/spaces/mteb/leaderboard}
}.
In our experiments we use the similarity threshold $\delta=0.5$\footnote{In our initial experiments $\delta=0.5$ yielded the best results.}.



\subsection{Datasets}
All of the configurations are executed on two judgmental forecasting datasets.
\paragraph{GJOpen}\footnote{\url{https://www.gjopen.com/}}
A forecasting arguments dataset \cite{Gorur2025Coherence}, which contains 2923 rephrased question-answer pairs from Good Judgment Open. Each pair was converted into a natural-language forecasting argument (claim) using the Mistral-7B-Instruct-v0.3 LLM~\cite{jiang2023mistral}, followed by manual review. The dataset covers both binary and multiple-choice forecasting questions, and is publicly available.

\paragraph{Metaculus}\footnote{\url{https://www.metaculus.com/}}
A subset of the Forecasting dataset from \cite{Halawi2024approaching}, which originally includes 8881 binary forecasting questions. We picked only the questions that have been resolved and were open between the dates of September 2023 and September 2024, which yielded 388 forecasting questions. We converted the questions into forecasting arguments (claims) using the Mistral-7B-Instruct-v0.3 LLM, followed by manual review, similarly to \cite{Gorur2025Coherence}.

\subsection{External Sources}
To support RAG, we retrieve 
evidence from two news article sites.
\paragraph{NYTimes}
We collected all NYTimes article abstracts published between January 2023 and December 2024 using the NYTimes API.

\paragraph{Guardian}
For each forecasting argument, we used GPT‑4o‑mini to generate five targeted search queries  
\ifthenelse{\boolean{arxivversion}}{(see Appendix~\ref{app:query_generation} for details)}{(see \cite{arxivversion} for details)}. To get article abstracts these queries were submitted to the Guardian API.

\paragraph{Retrieval Pipeline}
For both sources, to optimise evidence retrieval, we embedded each abstract using the Jina-V3 embedding model \cite{Sturua2025JinaV3}. The resulting embeddings were stored in a vector database, ChromaDB, to support real-time retrieval. We then retrieved the top five relevant articles for each claim, restricted to abstracts dated before the closing date of the claim.

\subsection{Models}
We evaluate all configurations using the following LLMs:
Mixtral (Mixtral-8x7B-Instruct-v0.1)~\cite{jiang2024mixtral}, Mistral (Mistral-7B-Instruct-v0.3)~\cite{jiang2023mistral}, Gemma (Gemma-7b-it)~\cite{gemmateam2024gemma}, Gemma-2 (Gemma-2-9b-it)~\cite{gemma2}, Llama-3 (Meta-Llama-3-8B-Instruct)~\cite{llama3}, and GPT-4o (GPT-4o-mini)~\cite{gpt-4o-mini} models. These models were chosen to ensure their training data cut-off dates preceded the forecasting events, thereby avoiding data contamination.


\section{Results}
\subsection{Individual Agents}

\paragraph{ArgLLM and RAG-ArgLLM Agents}
We first evaluate the ArgLLM and RAG-ArgLLM agents (using the NYTimes and the Guardian) with their four variations depth 1 (with 0.5 base score and estimated base score assigned to the claim) and depth 2 (with 0.5 base score and estimated base score assigned to the claim).
Table~\ref{tab:rag_argllm_d1_d2} (minus the last column) reports results of the ArgLLMs and the RAG-ArgLLMs agents, using the NYTimes and the Guardian, on the GJOpen and Metaculus datasets.

Retrieval improves forecasting accuracy on the Metaculus dataset in nearly all variations. The highest accuracy increase comes when the external sources are used with Gemma-2 (with depth 1, estimated base score), accuracy improves from 68\% to 81\% (for both NYTimes and Guardian). The RAG-ArgLLM agents that did not improve accuracy were already performing well and the external sources changed the argumentative reasoning of those models.
On GJOpen, the improvement is less consistent, however, using the Guardian seems to improve forecasting accuracy more compared to using the NYTimes.
The highest accuracy increase occurs when using the NYTimes with Llama-3 (with depth 2, 0.5 base score) where the accuracy improves from 63\% to 75\%.
We additionally analyse the overlap between supporting and attacking arguments generated by RAG-ArgLLM agents when different external sources are used.  
\ifthenelse{\boolean{arxivversion}}{Table~\ref{tab:similarity} in Appendix~\ref{app:overlap} shows that overlap is generally high}{Overall, we can observe that overlap is generally high (see \cite{arxivversion} for details)}, indicating that different sources often lead to substantially different argumentative structures.

\paragraph{RbAM Agents}
The last column in Table~\ref{tab:rag_argllm_d1_d2} shows the results of the RbAM agent with two external sources (the NYTimes and the Guardian) on the two forecasting datasets.

The RbAM agent does not seem to perform that well for the GJOpen dataset, this is because the base scores generated are very small due to the abstracts not having a proper argumentative structure.
One possible solution is to prompt a language model to extract or generate structured arguments from each abstract, which we leave to future work.
The Guardian as an external source helped more than the NYTimes on the GJOpen dataset. This is possibly because of its more relevant context.
On the other hand, the RbAM agent performs better for the Metaculus datasets, although it still does not surpass the best ArgLLM or RAG-ArgLLM agent.
Therefore, we decided not to combine QBAFs obtained from the RbAM agent as the accuracy was very low.

\begin{table}[htp!]
    \centering
    \footnotesize
    \setlength{\tabcolsep}{0.1em}
    \renewcommand{\arraystretch}{0.9}
    \begin{tabular}{lcl|cccccc|c}
         \toprule
         Dataset & Depth & Source & Mixtral & Mistral & Gemma & Gemma-2 & Llama-3 & GPT-4o & RbAM \\
         \midrule
         \multirow{6}{*}{GJOpen}
            & \multirow{3}{*}{1}
                & W/O & {\bf 67}/70 & 65/71 & {\bf 73}/57 & {\bf 79}/{\bf 77} & 78/75 & {\bf 74}/{\bf 69} & - \\
            &     & NYTimes & 61/71 & 65/73 & 71/{\bf 66} & 76/75 & 78/44 & 72/66 & 43/29 \\
            &     & Guardian & 61/{\bf 72} & {\bf 66}/{\bf 74} & 72/64 & 78/76 & {\bf 79}/{\bf 76} & 71/67 & {\bf 49}/{\bf 36} \\ 
            
            \cmidrule{2-10}
            
            & \multirow{3}{*}{2}
                & W/O &  {\bf 56}/{\bf 69} &  {\bf 56}/67 &  {\bf 70}/56 &  {\bf 79}/{\bf 77} & 63/{\bf 75} & 69/{\bf 68} & - \\
            &     & NYTimes & 49/67 & 52/{\bf 68} & 64/{\bf 63} & 75/75 &  {\bf 75}/50 & 71/66 & - \\
            &     & Guardian & 49/68 & 51/{\bf 68} & 64/{\bf 63} & 77/76 & 74/52 &  {\bf 72}/67 & - \\
         
         \midrule
         
         \multirow{6}{*}{Metaculus}
            & \multirow{3}{*}{1}
                & W/O & 66/74 & 61/70 & 76/73 & 68/66 & {\bf 81}/65 & {\bf 81}/74 & - \\
            &     & NYTimes & {\bf 71}/{\bf 78} & {\bf 79}/{\bf 80} & {\bf 81}/73 & {\bf 81}/{\bf 78} & {\bf 81}/{\bf 78} & 78/73 & {\bf 69}/{\bf 79} \\
            &     & Guardian & 69/73 & {\bf 79}/{\bf 80} & 80/{\bf 76} & {\bf 81}/{\bf 78} & 79/69 & 79/{\bf 76} & 63/78 \\ 
            
            \cmidrule{2-10}
            
            & \multirow{3}{*}{2}
                & W/O &  {\bf 61}/{\bf 73} & 62/67 & 71/73 & 67/66 &  {\bf 78}/65 &  {\bf 82}/{\bf 76} & - \\
            &     & NYTimes & 52/72 & 61/{\bf 78} &  {\bf 72}/{\bf 74} & 76/74 & {\bf 78}/66 & 79/73 & - \\
            &     & Guardian & 55/{\bf 73} &  {\bf 66}/77 & 68/73 &  {\bf 80}/{\bf 78} & 73/{\bf 68} & 80/{\bf 76} & - \\
         \bottomrule
    \end{tabular}
    \caption{Performance of individual agents on the GJOpen and Metaculus datasets, across Depth 1 and Depth 2. The table compares the baseline ArgLLM agents (W/O source) and the RAG-ArgLLM agents (NYTimes and Guardian source) with the base LLM models as per columns, as well as, in the last column, the RbAM agent (using the NYTimes and the Guardian, as per respective rows). For each agent configuration, we report accuracy of 0.5 base /estimated base scores. Bold indicates best results for
    each agent configuration, with/out the use of sources.}
    \label{tab:rag_argllm_d1_d2}
\end{table}


\subsection{Multi-Agent Claim Verification}
\paragraph{ArgLLM Agent Pairs}
We first evaluate our multi-agent framework across six pairs of ArgLLM agents with their four variations depth 1 (with 0.5 base score and estimated base score assigned to the claim) and depth 2 (with 0.5 base score and estimated base score assigned to the claim). For each variation, we apply two aggregation functions, average and maximum, to combine argument base scores.
\ifthenelse{\boolean{arxivversion}}
{Table~\ref{tab:combined} in Appendix~\ref{app:argllm_pairs} reports results for depth 1 and depth 2 for the GJOpen and Metaculus datasets.}{We evaluate depth 1 and depth 2 on the GJOpen and Metaculus datasets, full results are available in \cite{arxivversion}.}
Each ArgLLM agent independently generates arguments and their base scores, then our multi-agent framework combines pairs of ArgLLM agents.

The results show variation across individual ArgLLM agents.
This variation reflects differences in the agents' argumentative reasoning capabilities.
Across all variations, the combination of argumentative reasoning produced by the ArgLLM agents frequently outperforms at least one of the individual models, and in several cases outperform both.
Our 
framework improves performance when the paired agents bring complementary argumentative reasoning. 
For example, Table~\ref{tab:llama3_combined} shows Llama-3 combined with the other agent tend to improve on accuracy which could mean that Llama-3 generates argumentative reasoning that complements other agents.
The instances where the 
framework does not perform well are when a base agent already performs well individually. This could suggest that the second agent contributes less useful or redundant arguments. For instance, Gemma-2 already performs well, so combining it with other agents does not 
improve results.

Overall, the multi-agent framework can improve forecasting performance, particularly when the base agents offer diverse argumentative perspectives. Average aggregation generally outperforms max aggregation, suggesting that a moderate base score is more effective than selecting the strongest base score.

\begin{table}[htp!]
\centering
\setlength{\tabcolsep}{0.3em}
\footnotesize
\renewcommand{\arraystretch}{0.9}
\begin{tabular}{llrcc|ccc}
\toprule
 & \textbf{Dataset} & \textbf{Model} & \multicolumn{2}{c}{\textbf{Llama-3 (D=1)}} &  & \multicolumn{2}{c}{\textbf{Llama-3 (D=2)}} \\
\cmidrule(lr){4-5} \cmidrule(lr){7-8}
& & & Avg & Max &  & Avg & Max \\
\midrule

\multirow{12}{*}{\textbf{ArgLLM}} 
& \multirow{5}{*}{GJOpen} 
    & Single & \multicolumn{2}{c}{78/75} & Single & \multicolumn{2}{c}{63/75} \\
    & & Mixtral 67/70 & 68/\underline{75} & 77/\underline{75} & 56/69 & 43/74 & 49/\underline{75} \\
    & & Mistral 65/71 & 69/\underline{75} & 75/\underline{75} & 56/67 & 49/\underline{75} & 49/\underline{75} \\
    & & Gemma 73/57 & 73/\textbf{76} & \underline{78}/\textbf{78} & 50/70 & 70/56 & 50/70 \\
    & & Gemma-2 79/77 & 78/\textbf{78} & 78/\textbf{78} & 79/77 & 50/70 & 50/70 \\
\cmidrule(lr){2-8}
& \multirow{5}{*}{Metaculus} 
    & Single & \multicolumn{2}{c}{81/65} & Single & \multicolumn{2}{c}{81/65} \\
    & & Mixtral 66/74 & 66/\underline{74} & 66/\underline{74} & 61/73 & 66/{\bf 76} & 66/{\bf 76} \\
    & & Mistral 61/70 & 61/\underline{70} & 61/\underline{70} & 62/67 & 61/{\bf 74} & 61/{\bf 74} \\
    & & Gemma 76/73 & 68/66 & 76/\underline{73} & 76/\underline{73} & 71/73 & 68/66 \\
    & & Gemma-2 68/66 & 68/\underline{66} & 68/\underline{66} & 67/66 & 68/\underline{66} & 68/\underline{66} \\
\midrule

\multirow{12}{2pt}{\textbf{RAG-ArgLLM NYTimes}} 
& \multirow{5}{*}{GJOpen} 
    & Single & \multicolumn{2}{c}{78/44} & Single & \multicolumn{2}{c}{75/50} \\
    & & Mixtral 61/71 & 61/{\bf 71} & 61/\underline{71} & 49/67 & 37/49 & 48/51 \\
    & & Mistral 59/68 & 65/{\bf 73} & 65/{\bf 73} & 52/68 & 41/49 & 48/50 \\
    & & Gemma 71/66 & 76/{\bf 75} & 76/{\bf 75} & 64/63 & 53/54 & 37/57 \\
    & & Gemma-2 76/75 & 76/\underline{75} & 37/57 & 75/75 & 66/71 & 76/\underline{75} \\
\cmidrule(lr){2-8}
& \multirow{5}{*}{Metaculus} 
    & Single & \multicolumn{2}{c}{79/68} & Single & \multicolumn{2}{c}{78/66} \\
    & & Mixtral 71/78 & 71/\underline{78} & 71/\underline{78} & 52/72 & 41/63 & 52/66 \\
    & & Mistral 79/80 & \underline{79}/\underline{80} & \underline{79}/\underline{80} & 61/78 & 53/67 & 56/65 \\
    & & Gemma 81/73 & \underline{81}/78 & \underline{81}/\underline{73} & 72/74 & 64/70 & 44/47 \\
    & & Gemma-2 81/78 & \underline{81}/78 & \underline{81}/\underline{78} & 76/74 & 71/70 & 44/47 \\
\midrule

\multirow{12}{2pt}{\textbf{RAG-ArgLLM Guardian}} 
& \multirow{5}{*}{GJOpen} 
    & Single & \multicolumn{2}{c}{78/76} & Single & \multicolumn{2}{c}{74/52} \\
    & & Mixtral 61/72 & 61/72 & 61/72 & 46/68 & 36/50 & 51/52 \\
    & & Mistral 66/74 & 66/74 & 66/74 & 51/68 & 39/51 & 48/51 \\
    & & Gemma 72/64 & \underline{78}/\underline{76} & \underline{78}/\underline{76} & 64/63 & 52/56 & 38/59 \\
    & & Gemma-2 78/76 & \underline{78}/\underline{76} & \underline{78}/\underline{76} & 77/76 & 71/73 & 38/59 \\
\cmidrule(lr){2-8}
& \multirow{5}{*}{Metaculus} 
    & Single & \multicolumn{2}{c}{79/70} & Single & \multicolumn{2}{c}{73/68} \\
    & & Mixtral 69/73 & 69/\underline{73} & 69/\underline{73} & 55/73 & 45/64 & 55/66 \\
    & & Mistral 79/80 & \underline{79}/\underline{80} & \underline{79}/\underline{80} & 66/77 & 56/66 & 61/66 \\
    & & Gemma 80/76 & {\bf 81}/{\bf 78} & \underline{80}/\underline{76} & 68/73 & 62/69 & 45/48 \\
    & & Gemma-2 81/78 & \underline{81}/\underline{78} & \underline{81}/\underline{78} & 80/78 & 76/75 & 45/48 \\
\bottomrule
\end{tabular}
\caption{
Accuracy results of our multi-agent combination framework applied to 
pairs of 
(Llama-3 and one of Mixtral, Mistral, Gemma and Gemma-2)
(i) ArgLLM agents; (ii) RAG-ArgLLM agents using NYTimes; (iii) RAG-ArgLLM agents using the Guardian, for Depth 1 (left) and Depth 2 (right).
We show accuracy results using 
0.5/estimated base scores and $\omega_{avg}$ (Avg) and $\omega_{max}$ (Max) base score aggregations. 
Bold indicates improvement over both 
Single agents; underline indicates improvement over one and parity with the other.}
\label{tab:llama3_combined}
\end{table}

\paragraph{RAG-ArgLLM Agent Pairs}
Next, we evaluate our multi-agent framework across six pairs of RAG-ArgLLM agents with their four variations depth 1 (with 0.5 base score and estimated base score assigned to the claim) and depth 2 (with 0.5 base score and estimated base score assigned to the claim) and the NYTimes and the Guardian as external sources.
\ifthenelse{\boolean{arxivversion}}
{Tables~\ref{tab:combined_nytimes} and~\ref{tab:combined_guardian} in Appendix~\ref{app:ragargllm_pairs} report results for the multi-agent framework using the RAG-ArgLLM agents with the NYTimes and the Guardian (respectively) for the GJOpen and Metaculus datasets.}{Results for the multi-agent framework using RAG-ArgLLM agents with NYTimes and the Guardian on the GJOpen and Metaculus datasets are provided in \cite{arxivversion}.}

Results show that multi-agent framework with the RAG-ArgLLM agents often outperforms at least one of the individual RAG-ArgLLM agents, and in some cases exceed both. This similarly confirms combining QBAFs improve performance when the paired agents bring complementary argumentative reasoning, even when arguments are retrieved from the external sources. Notably, looking at Table~\ref{tab:llama3_combined}, combinations involving Llama-3 often improves accuracy despite Llama-3 not performing well individually with the RAG-ArgLLM agents, indicating its arguments remain complementary when combined with other models.
We also observe that average aggregation typically outperforms max, aligning with earlier findings for the multi-agent framework.
Performance gains are more consistent on the Metaculus dataset, where the paired agents often match or exceed both individual models.
On the other hand, improvements on the GJOpen dataset are more dependent on the choice of the external source. The Guardian tends to be more effective than the NYTimes, consistent with the findings for the single-agent RAG-ArgLLM and RbAM agents.

Overall, our multi-agent framework applied to the RAG-ArgLLM agent pairs 
shows that combining retrieval-augmented models can also be beneficial, particularly when the combination is done with average aggregation and when the base models offer diverse argumentative perspectives.

\paragraph{Two-Agent Combination}
We then explore the performance of combining two different agents, analysing three variants: combining two RAG-ArgLLM agents that use different sources (the NYTimes and the Guardian), and combining standard ArgLLM agents with the RAG-ArgLLM agents (using either the NYTimes or the Guardian).

\ifthenelse{\boolean{arxivversion}}
{Tables~\ref{tab:combined_nytimes_guardian_d1} and \ref{tab:combined_nytimes_guardian_d2} in Appendix~\ref{app:two-agent} show the results of combining two RAG-ArgLLM agents, where one uses the NYTimes and the other uses the Guardian as external sources.}{Results for combining two RAG-ArgLLM agents, where one uses the NYTimes and the other uses the Guardian as external sources, are provided in \cite{arxivversion}.}
The combination of the RAG-ArgLLM agents with different sources often yields significant improvements, particularly in depth 1.
Mistral (Guardian) also proves to be a consistently strong complementary agent in Depth 1, improving or maintaining accuracy in most pairings across both datasets (five out of six in both).
However, performance in depth 2 is less consistent and often leads to a degradation in accuracy. This is likely due to the generation of irrelevant arguments in the second layer. Nonetheless, even in depth 2 some improvement is still observable over single agents.

\ifthenelse{\boolean{arxivversion}}
{Tables~\ref{tab:combined_nytimes_argllm_d1} and \ref{tab:combined_nytimes_argllm_d2} in Appendix~\ref{app:two-agent} present results where the ArgLLM agents are paired with the RAG-ArgLLM agents using the NYTimes. Tables~\ref{tab:combined_guardian_argllm_d1} and \ref{tab:combined_guardian_argllm_d2} in Appendix~\ref{app:two-agent} show similar pairings but with the Guardian as the external source.}{Results for pairings of ArgLLM and RAG-ArgLLM agents using NYTimes or Guardian external sources are provided in \cite{arxivversion}.}
The results indicate that this hybrid approach is highly effective. The RAG-ArgLLM agents introduce externally grounded arguments that can complement the internal reasoning of the ArgLLM agents.
The benefit is particularly clear when using the NYTimes
. For example, Mistral (NYTimes) often improved or maintained the accuracy of its ArgLLM partner on the GJOpen dataset (five out of six instances) for average aggregation.
The Guardian source also shows strong performance in specific pairings.
However, the RAG-ArgLLM agents using the Guardian does not seem to improve accuracy when it is combined with the ArgLLM agents as much.

Overall, our multi-agent framework applied to two different agents demonstrates that combining different external sources can be beneficial. The most significant gains are achieved when agents leverage different external sources.

\paragraph{Three-Agent Combination}
We extend the analysis of our multi-agent framework to three agents, where a standard ArgLLM is combined with two RAG-ArgLLM agents, one using the NYTimes and the other the Guardian.
\ifthenelse{\boolean{arxivversion}}
{The results, presented in Tables~\ref{tab:d1_argllm_mixtral_nytimes_guardian}-\ref{tab:d2_argllm_gpt4o_nytimes_guardian} in Appendix~\ref{app:three-agent}, demonstrate the potential to increase accuracy by increasing the diversity of argumentative reasoning.}{The results demonstrate the potential to increase accuracy by increasing the diversity of argumentative reasoning, detailed results are provided in \cite{arxivversion}.}

The addition of a third agent often leads to more robust and accurate forecasts, particularly on the GJOpen dataset.
The most consistent improvements are observed when the base ArgLLM agent is a strong performer itself, such as with Gemma-2 or GPT-4o. This suggests that when an agent's internal reasoning is strong, it provides a solid foundation that can be effectively enhanced by complementary information from external sources, thereby improving the forecasting accuracy. While the three-agent combination does not always outperform the best single agent in the trio, it consistently shows a clear improvement over the other two individual agents.
On the Metaculus dataset, the benefits are also apparent, although the strong individual agent performance sometimes sets a high bar for improvement.
Both aggregation methods show instances of improvement. However, average aggregation performs better than max aggregation in most instances, supporting the evidence from previous experiments.

Overall, the three-agent framework, by increasing the diversity of complementary reasoning, can improve forecasting accuracy whether from different models or different external data sources.


\section{Conclusions and Future Work}
\label{sec:conclusions}
In this paper, we introduce a novel multi-agent framework which integrates multiple perspectives by combining QBAFs generated by diverse LLM-based agents.
We designed and evaluated three distinct agent types: the baseline ArgLLM agents (which relies on internal LLM knowledge), the RAG-ArgLLM agents and the RbAM agents (which ground their reasoning in external sources).

Our empirical evaluation on the GJOpen and Metaculus datasets yielded several key insights. First, the RAG-ArgLLM agents consistently outperformed the ArgLLM agents, especially on the Metaculus dataset. This confirms that grounding argumentative reasoning with external evidence can improve forecasting accuracy. While the RbAM agents showed promise on the Metaculus dataset, they did not perform well on the GJOpen dataset.
Second, our multi-agent framework applied to pairs of the same agent types, in some instances improved forecasting accuracy. Performance gains were most significant when the agents generated complementary argumentative perspectives.
Pairing an ArgLLM agent with a RAG-ArgLLM agent, or combining two RAG-ArgLLM agents using different external sources (the NYTimes and the Guardian), proved highly effective. This suggests integrating both internal LLM reasoning and external evidence can be beneficial.
Finally, instantiating our framework with three agents, an ArgLLM and two RAG-ArgLLMs with different sources, yielded further improvements. By increasing the diversity of reasoning and external sources, the three-agent framework produced more robust reasoning. This suggests that combining more complementary perspectives yields better results.


Building on these findings, we plan to explore several promising directions: 
(1) investigating in more depth which models to pick while combining their argumentative reasoning; 
(2) studying whether an interactive debate could improve forecasting accuracy (e.g. using a similar framework to \citet{Rago2023AX});
(3) experimenting with a larger number of frameworks; (4) evaluating the framework's robustness against real-world data irregularities such as noisy/misleading data, biased sources, and a wider variety of question types; and (5) investigating the practical deployment of the system, including optimising the computational cost and addressing the integration challenges into existing workflows.


\begin{acks}
This research was partially funded by the ERC under the EU’s Horizon 2020 research and innovation programme (grant agreement no. 101020934, ADIX), by J.P. Morgan and by the Royal Academy of Engineering, UK, under the Research Chairs and Senior Research Fellowships scheme (grant agreement no. RCSRF2021\textbackslash 11\textbackslash 45).
\end{acks}



\bibliographystyle{ACM-Reference-Format} 
\bibliography{bibliography}


\ifthenelse{\boolean{arxivversion}}{
\clearpage
\section*{Appendix}

\section*{Proofs}
\label{app:proof}

\begin{proposition}
    Both $\omega_{\text{avg}}$ and $\omega_{\max}$ satisfy Definition~\ref{def:aggregation}.
\end{proposition}

\begin{proof}
\label{proof:aggregation}
To prove $\omega_{\text{avg}}$ and $\omega_{\max}$ satisfy Definition~\ref{def:aggregation} we fix $k\in\{1,\dots,K\}$ and $v=(v_1,\dots,v_k)\in[0,1]^k$.

\textbf{Order-independence}
For $\omega_{\text{avg}}$, the sum $\sum_{i=1}^k v_i$ is invariant under permutations, hence so is $\frac{1}{k}\sum_i v_i$.
For $\omega_{\max}$, the set $\{v_1, \dots, v_k\}$ is unchanged by permutation, so its maximum is unchanged.

\textbf{Boundedness}
Let $m=\min(v)$ and $M=\max(v)$.
For $\omega_{\text{avg}}$, since $\min(v) \leq v_i \leq \max(v)$ for all $i$,
$k\min(v) \leq \sum_{i=1}^k v_i \leq k\max(v)$, then $\min(v) \leq \frac{1}{k}\sum_{i=1}^k v_i \leq \max(v)$.
For $\omega_{\max}$, $\omega_{\max}(v)=\max(v)$, so $\min(v) \leq \omega_{\max}(v)=\max(v) \leq \max(v)$.

\textbf{Idempotence}
If $v=(v_i, \dots, v_i)$, then $\omega_{\text{avg}}(v)=\frac{1}{k}\cdot k v_i=v_i$ and
$\omega_{\max}(v)=\max\{v_i, \dots, v_i\}=v_i$.

\textbf{Monotonicity}
Let $v'=(v_1', \dots, v_k')$ with $v_i \leq v_i'$ for all $i$.
For $\omega_{\text{avg}}$,
$\sum_{i=1}^k v_i \le \sum_{i=1}^k v_i'$, then $\frac{1}{k}\sum_{i=1}^k v_i \leq \frac{1}{k}\sum_{i=1}^k v_i'$. Thus $\omega_{\text{avg}}(v) \le \omega_{\text{avg}}(v')$.

For $\omega_{\max}$, since each $v_i \leq v_i'$, let $i=\arg\max(v)$ (i.e.\ $v_i=\max(v)$), by assumption $v_i \leq v_i'$ hence $\max(v)=v_i \leq v_i' \leq \max(v')$. Thus $\omega_{\max}(v) \leq \omega_{\max}(v')$.

All four properties hold for both aggregation functions.
\end{proof}


\lifting*

\begin{proof}
Let us first prove the case for $\mathcal{A}^*$.

($\Rightarrow$) 
By construction of $\mathcal{A}^*$ from Definition~\ref{def:combine}, since $(x^*, y^*) \in \mathcal{A}^*$, it must be the case that $\exists (x, y) \in \mathcal{A}$, where $x \in x^*$ and $y \in y^*$. 
We know from the same definition that $\mathcal{A}=\bigcup_{j=1}^n \mathcal{A}_j$ 
, and so it must be the case that $(x, y) \in \mathcal{A}_i$ for some $i \in \{1,\dots,n\}$.

($\Leftarrow$) 
We know that $(x, y) \in \mathcal{A}_i$, for some $i \in \{1,\dots,n\}$, and so, according to property (1) in Definition~\ref{def:combine}, $\exists_1 x^* \in \mathcal{X}^*$ such that $x \in x^*$, and $\exists_1 y^* \in \mathcal{X}^*$ such that $y \in y^*$. Then, also by Definition~\ref{def:combine}, $(x^*, y^*) \in \mathcal{A}^*$ if $(x, y) \in \mathcal{A}$, which we know is the case, as $\mathcal{A}=\bigcup_{j=1}^n \mathcal{A}_j$ and so $\mathcal{A}_i \subseteq \mathcal{A}$
.

The proof similarly follows for $\mathcal{S}^*$.
\end{proof}

\qbaf*
\begin{proof}
    By construction (from Definition~\ref{def:combine}), $\mathcal{X}^* \subseteq 2^{\mathcal{X}}$ where $\mathcal{X}=\bigcup_{i=1}^n \mathcal{X}_i$. Each $x \in \mathcal{X}$ belongs to exactly one cluster $x \in x^* \in \mathcal{X}^*$ (property 1 from Definition~\ref{def:combine}). Hence, $\mathcal{X}^*$ is a set of arguments.
    Thus, by construction of $\mathcal{A}^*$ from Definition~\ref{def:combine}, $\mathcal{A}^* \subseteq \mathcal{X}^* \times \mathcal{X}^*$. 
    The proof follows similarly for $\mathcal{S}^* \subseteq \mathcal{X}^* \times \mathcal{X}^*$.
    Let us prove by contradiction that $\mathcal{A}^* \cap \mathcal{S}^* = \emptyset$. Assume that $\exists (x^*, y^*) \in \mathcal{A}^* \cap \mathcal{S}^*$. Then, by Lemma \ref{lemma:lifting}, $\exists_{x, y \in \mathcal{X}} (x, y) \in \mathcal{A} \cap 
    \mathcal{S}$. 
    This cannot be the case as it contradicts the lingua-franca assumption that no ordered pair is an attack in one QBAF and a support in another. Hence, we have the contradiction and $\mathcal{A}^* \cap \mathcal{S}^*=\emptyset$.
    For each cluster $x^*\in\mathcal{X}^*$, $\tau^*(x^*)=\omega(\tau(x_1), \ldots, \tau(x_k))$ with $x_i \in \mathcal{X}^*$, by Definition~\ref{def:combine}. Since, $\omega$ maps to $[0,1]$, by Definition~\ref{def:aggregation}, $\tau^*: \mathcal{X}^* \rightarrow [0, 1]$ is a total function.
    Therefore, $\mathcal{Q}^*=\langle\mathcal{X}^*, \mathcal{A}^*, \mathcal{S}^*, \tau^*\rangle$ satisfies the definition of a QBAF.
\end{proof}

\qbafforf*
\begin{proof}
        We know from Proposition~\ref{prop:qbaf} that $\mathcal{Q}^*$ is a QBAF, thus we only need to show that $\mathcal{Q}^*$ is a QBAF for $\{ f \}$, so we take the three conditions for this in turn. \\
        (i) Let us prove by contradiction that $\forall_{x^* \in \mathcal{X}^*} \Path(\{f\}, x)\!=\!\emptyset$.
        Assume that $\exists_{x^* \in \mathcal{X}^*} \Path(\{f\}, x^*) \neq \emptyset$.
        Then, it must be the case that $\exists_{x^* \in \mathcal{X}^*} (\{f\}, x^*) \in \mathcal{A}^* \cup \mathcal{S}^*$. Then, by Lemma \ref{lemma:lifting}, $\exists_{x \in \mathcal{X}} (f, x) \in \mathcal{A} \cup \mathcal{S}$.
        However, we know that this cannot be the case as $\forall_{x \in \mathcal{X}} \Path(f, x)=\emptyset$ (from property (i) of the definition of a QBAF for $f$) 
        .
        Hence, we have the contradiction and $\forall_{x^* \in \mathcal{X}^*} \Path(\{f\}, x)\!=\!\emptyset$. \\
        (ii) Let us show that $\forall_{x^* \in \mathcal{X}^* \setminus \{f\}} \mid \Path(x^*, \{f\}) \mid = 1$.
        We know that $\forall_{x \in \mathcal{X} \setminus f} \mid \Path(x, f) \mid = 1$ (from property (ii) of the definition of a QBAF for $f$), where $\Path(x, f)=\langle (x, x_1), \ldots, (x_k, f) \rangle$. Each relation in this path is $(x, x_1), (x_i, x_{i + 1}), (x_k, f) \in \mathcal{A} \cup \mathcal{S}$.
        Then, by Lemma~\ref{lemma:lifting}, $(x^*, x_1^*), (x_i^*, x_{i + 1}^*), (x_k^*, \{f\}) \in \mathcal{A} \cup \mathcal{S}$, which creates a unique path $\Path(x^*, \{f\})=\langle (x^*, x_1^*), \ldots, (x_k^*, \{f\}) \rangle$. 
        Therefore, $\forall_{x^* \in \mathcal{X}^* \setminus \{f\}} \mid \Path(x^*, \{f\}) \mid = 1$.
        \\
        (iii) 
        Let us prove by contradiction that $\forall_{x^* \in \mathcal{X}^*} \Path(x^*, x^*)=\emptyset$. Assume that $\exists_{x^* \in \mathcal{X}^*} \Path(x^*, x^*) \neq \emptyset$.
        Let this path be $\langle (x^*, x_1^*), \ \ldots, \\ (x_k^*, x^*)\rangle$. By Lemma~\ref{lemma:lifting}, each relation $(x_i^*, x_{i+1}^*)$ in this cycle corresponds to an underlying relation $(x_{i}, x_{i + 1}) \in \mathcal{A} \cup \mathcal{S}$, where $x_{i} \in x_i^*$ and $x_{i + 1} \in x_{i + 1}^*$
        , giving a path $\langle (x, x_1), \ldots, (x_k, x)\rangle$, where $x, x_1, x_k \in \mathcal{X}_i$. However, $\mathcal{Q}_i$ cannot have cycles (from property (iii) of a QBAF for $f$). This contradicts our assumption. Therefore, no cycles can exist in $\mathcal{Q}^*$.
    
    All three properties hold, therefore $\mathcal{Q}^*$ is a QBAF for $\{f\}$.
\end{proof}

\qbafprocon*
\begin{proof}
Let us first prove the case for pro-arguments.

We know from Proposition~\ref{prop:qbaf_for_f} that $\mathcal{Q}^*$ is a QBAF for $\{f\}$. \\
($\Rightarrow$) We know that $x^* \in \Pro(\mathcal{Q}^*)$.
By definition of pro-arguments, the unique path $p^* = \Path(x^*, \{f\}) = \langle (x^*, y^*), \ldots, (z^*, \{f\})$ has an even number of attacks, i.e. $\mid p^* \cap \mathcal{A}^* \mid$ is even. Each relation in this path is $(x^*, x_1^*), (x_j^*, x_{j + 1}^*), (x_k^*, \{f\}) \in \mathcal{A}^* \cup \mathcal{S}^*$.
By Lemma~\ref{lemma:lifting}, each relation $(x^*, x_1^*), (x_j^*, x_{j + 1}^*), (x_k^*, \{f\})$ in this path corresponds to an underlying relation $(x, x_1), (x_{j}, x_{j + 1}), (x_k, f) \in \mathcal{A} \cup \mathcal{S}$, which creates a path $p = \Path(x, f)=\langle (x, x_1), \ldots, (x_k, f) \rangle$.
Thus, $\mid p^* \cap \mathcal{A}^* \mid = \mid p \cap \mathcal{A}_i \mid$. Since $\mid p^* \cap \mathcal{A}^* \mid$ is even, $\mid p \cap \mathcal{A}_i \mid$ must also be even. By definition of pro-arguments, this means $x \in \Pro(\mathcal{Q}_i)$.

($\Leftarrow$) We know that $\exists_{x \in x^*} x \in \Pro(\mathcal{Q}_i)$, for some $\mathcal{Q}_i$.
By the definition of pro-arguments, this means the path $p = \Path(x, f) = \langle (x, x_1), \ldots, (x_k, f) \rangle$ in $\mathcal{Q}_i$ has an even number of attacks, i.e. $\mid p \cap \mathcal{A}_i \mid$ is even.
Each relation in this path is $(x, x_1), (x_{j}, x_{j + 1}), (x_k, f) \in \mathcal{A} \cup \mathcal{S}$.
By Lemma~\ref{lemma:lifting}, each relation $(x, x_1), (x_{j}, x_{j + 1}), (x_k, f)$ in this path corresponds to a relation $(x^*, x_1^*), (x_j^*, x_{j + 1}^*), (x_k^*, \{f\}) \in \mathcal{A}^* \cup \mathcal{S}^*$, which creates a path $p^* = \Path(x^*, \{f\}) = \langle (x^*, y^*), \ldots, (z^*, \{f\})$.
Thus, $\mid p^* \cap \mathcal{A}^* \mid = \mid p \cap \mathcal{A}_i \mid$. Since $\mid p \cap \mathcal{A}_i \mid$ is even, $\mid p^* \cap \mathcal{A}^* \mid$ must also be even. By definition, this means $x \in \Pro(\mathcal{Q}^*)$.

The same logic applies to con arguments, where the number of attacks is odd. Thus the combined QBAF preserves the set of pro/con arguments of the original argument within the clustered arguments.
\end{proof}

\algorithm*
\begin{proof}
\label{proof:algorithm}
To prove that Algorithm~\ref{alg:cap} terminates we need to show that all the loops are finite. Main loop (Line 3) iterates up to the finite maximum depth of the input QBAFs. First inner loop (Line 5) iterates through a finite number of arguments in the previous depth. Second inner loop (Line 6) iterates through all pairs of finite arguments in the current depth. Third inner loop (Line 15) iterates through a finite number of arguments in the previous depth and the merged clusters. All inner loops iterate over finite sets. Therefore, the algorithm is guaranteed to terminate.

The proof is by double induction. The outer induction is on the number of QBAFs, $n$. The inner induction is on the depth of QBAFs, $d$.

\subsection*{(Outer Induction) Base case: $n=2$}
We show that the algorithm correctly produces a combined QBAF given two QBAFs $\mathcal{Q}_1$ and $\mathcal{Q}_2$. This is proven using an inner induction on depth $d$.

    \paragraph{(Inner Induction) Base case: $d=0$}
    Lines 1 initialises the combined QBAF where $\mathcal{X}^*$ is initialised with singleton clusters for all arguments in $x \in \mathcal{X}$, and only setting the base score of $a$ to $\tau^*(a^*) = \omega(\tau_1(a), \tau_2(a))$. This correctly implements the base score definition for the root. Thus, the proposition holds for $d=0$.

    \paragraph{(Inner Induction) Inductive Hypothesis}
    Assume that for all depths $d' < k$ (where $k \geq 1$), the algorithm has correctly constructed the combined QBAF, and all properties from Definition~\ref{def:combine} hold for arguments and relations up to depth $k-1$.

    \paragraph{(Inner Induction) Inductive Step: $d=k$}
    We show that iteration $k$ of the main loop (Line 5) correctly constructs the combined QBAF for depth $k$.
    \begin{enumerate}
        \item Correct construction of $\mathcal{X}^*$: Line 6 identifies all arguments at depth $k$, and Lines 7-12 partitions them into clusters.
        \begin{itemize}
            \item The algorithm initialises $\{x\} \in \mathcal{X}^*$ with singleton clusters for all arguments in $x \in \mathcal{X}$. The merging process on Line 8 combines existing clusters. Since an argument $x$ starts in a unique singleton cluster $\{x\}$ and clusters are only ever merged, $x$ will always belong to exactly one cluster, thus satisfying property (1) of Definition~\ref{def:combine}.
            
            \item ($\Rightarrow$) Assume $x,y$ are added to the same cluster $x^*$ (Line 8), then we know there exists $z, z' \in z*\in \mathcal{X}*$ with $(x,z), (y,z') \in \mathcal{A}$ or $(x,z), (y,z') \in \mathcal{S}$ (Line 6) and $\Psi(x,y)\geq\delta$ (Line 7).
            ($\Leftarrow$) Now assume we have $z, z' \in z^*\in \mathcal{X}^*$ with $(x,z), (y,z') \in \mathcal{A}$ or $(x,z), (y,z') \in \mathcal{S}$, and $\Psi(x,y)\geq\delta$. These conditions lead to the execution of adding $x,y$ to the same cluster $x^*$ (Line 8).
            
            
        \end{itemize}
        
        \item ($\Rightarrow$) Assume the algorithm (Line 16) adds $(x^*, z^*)$ to $\mathcal{A}^*$, which is executed if and only if the condition (Line 13), $\exists x \in x^*, \exists z \in z^*$ such that $(x,z) \in \mathcal{A}$, is true. This directly matches one direction of the definition.
        ($\Leftarrow$) Now assume there exists an attack $(x, z)$ where $x \in x^*$ and $z \in z^*$, with $x^*$ being a new cluster at depth $k$ and $z^*$ a cluster at depth $k-1$. Since the algorithm iterates through all such new clusters $x^*$ and all previous arguments $z$ (Line 13), the condition (Line 16) is evaluated for the specific pair $(x,z)$ and found to be true. Consequently, the relation $(x^*, z^*)$ will be added to $\mathcal{A}^*$.        
        Thus, the algorithm constructs $\mathcal{S}^*$ exactly as defined.
        
        \item Similar considerations apply to the support relations, $\mathcal{S}^*$, w.r.t. Lines 13 and 17-18.
        
        \item Assume given an arbitrary cluster $x^*\in\mathcal{X}^*$ created at iteration $k$. The algorithm (Line 14) assigns the base score to the cluster $x^*$ as $\tau^*(x^*)=\omega(\tau(x)\mid x\in x^*)$. This exactly matches the property in Definition~\ref{def:combine}.
    \end{enumerate}
    By the principle of induction on $d$, the algorithm is correct for all depths when $n=2$.

\subsection*{(Outer Induction) Inductive Hypothesis}
Assume Algorithm~\ref{alg:cap} has correctly constructed the combined QBAF for $n=m$ QBAFs, where $m \geq 2$, and all properties from Definition~\ref{def:combine} are satisfied.

\subsection*{(Outer Induction) Inductive Step: $n=m+1$}
We show the algorithm is correct for $n=m+1$ QBAFs. The proof follows the same inner induction on depth $d$. The algorithm's output is independent of the value of $n$; it operates on the union of all arguments. Since, set union is order independent (satisfies associativity) the reasoning for each step of the inner induction holds identically for $n=m+1$ QBAFs as it does for $n=2$. 
Thus, by the principle of double induction, the proposition is proven for all $n > 1$.

To show that the algorithm executes in polynomial time, let $n$ be the number of QBAFs, $N=\mid\mathcal{X}\mid$ be the total number of arguments across all frameworks, and $D$ be the maximum depth of any input QBAF.
The main loop (Lines 3-19) runs $D$ times. Inside the main loop, the second loop (Lines 5-12) iterates over $z$ in `previous layer', where `previous layer' is at most $N$. Inside the second loop, the third loop (Lines 6-12) iterates over all pairs $(x,y)$ that are related to $z$. In the worst case, a single node could be the parent of all arguments, which gives time complexity $O(N^2)$.
So, the total complexity for the second loop is $O(N^3)$.
The last loop (Lines 13-18) iterates over $z$ in `previous layer' and $x^*$ in `merged', which could be at most $N$ if all clusters are singletons.
The second loop and the last loop, each iteration of the main loop over $D$ takes at most $O(N^3)=O(N^3)+O(N^2)$ time. Since the main loop iterates $D$ times, the total complexity is $O(D\times N^3)$. Thus the algorithm executes in polynomial time.
\end{proof}

\section*{Query Generation for the Guardian API}
\label{app:query_generation}
Here, we explain the query generation step to be passed to the Guardian API to retrieve the relevant articles for the forecasting claims.

In order to generate queries we use the following prompt:

\noindent\rule{\linewidth}{0.4pt}
\begin{quote}
You are a **Query Generator Agent**. Your task is to convert a claim into **5 high-quality search queries**, using Boolean AND/OR, synonyms, and phrase quotes. **Do not use numeric range operators like 550000..800000 or natural language comparisons (greater than, less than).**

Example:
Input: Claim: 'UK inflation will go up by the end of 2025'
Output:
1. ("UK inflation" OR "UK consumer price index") AND ("end of 2025" OR "late 2025") AND (forecast OR outlook)
2. ("Bank of England" AND interest rate AND inflation AND ("Q4 2025" OR "end‑2025"))
3. ("UK CPI" AND ("price growth" OR inflation) AND ("Dec 2025" OR "2025 forecast"))

---

Input: Claim: \{claim\}
\end{quote}
\noindent\rule{\linewidth}{0.4pt}

Each forecasting claim is passed into the prompt to get up to five queries.

\section{Overlap of RAG-ArgLLM with Different Sources}
\label{app:overlap}
Table~\ref{tab:similarity} reports the percentage of overlapping arguments between agents using the NYTimes and the Guardian, across datasets and base models.

\begin{table*}[bt]
    \centering
    \begin{tabular}{crccccc}
        & & Mistral & Gemma & Gemma-2 & Llama-3 & GPT-4o-Mini \\
        \midrule

        \multirow{5}{*}{GJOpen}
        & Mixtral & 91/96 & 29/34 & 29/34 & 94/96 & 91/92 \\
        & Mistral & - & 36/35 & 36/35 & 89/96 & 87/92 \\
        & Gemma & - & - & 0/0 & 22/30 & 58/73 \\
        & Gemma-2 & - & - & - & 22/30 & 26/34 \\
        & Llama-3 & - & - & - & - & 26/34 \\

        \midrule
        \multirow{5}{*}{Metaculus}
        & Mixtral & 25/69 & 3/1 & 3/1 & 22/33 & 39/53 \\
        & Mistral & - & 74/26 & 74/26 & 71/48 & 67/56 \\
        & Gemma & - & - & 94/82 & 77/67 & 61/45 \\
        & Gemma-2 & - & - & - & 78/65 & 61/45 \\
        & Llama-3 & - & - & - & - & 58/52
    \end{tabular}
    \caption{Percentage of overlapping supporting/attacking arguments generated by RAG-ArgLLM agents using NYTimes (column) and Guardian (row).}
    \label{tab:similarity}
\end{table*}

\section{Multi-Agent Claim Verification}
\subsection{ArgLLM Agent Pairs}
\label{app:argllm_pairs}
Table~\ref{tab:combined} reports results for depth 1 and depth 2 for the GJOpen and Metaculus datasets.

\begin{table*}[bt]
    \centering
    \footnotesize
    \setlength{\tabcolsep}{0.28em}
    \begin{tabular}{crcccccccccc|crcccccccccc}
        & & \multicolumn{2}{c}{Mistral} & \multicolumn{2}{c}{Gemma} & \multicolumn{2}{c}{Gemma-2} & \multicolumn{2}{c}{Llama-3} & \multicolumn{2}{c|}{GPT-4o}
        & & \multicolumn{2}{c}{Mistral} & \multicolumn{2}{c}{Gemma} & \multicolumn{2}{c}{Gemma-2} & \multicolumn{2}{c}{Llama-3} & \multicolumn{2}{c}{GPT-4o} \\
        & & Avg & Max & Avg & Max & Avg & Max & Avg & Max & Avg & Max
        & & Avg & Max & Avg & Max & Avg & Max & Avg & Max & Avg & Max \\
        \midrule

        \multirow{7}{*}{GJOpen}
        & Single & \multicolumn{2}{c}{65/71} & \multicolumn{2}{c}{73/57} & \multicolumn{2}{c}{79/77} & \multicolumn{2}{c}{78/75} & \multicolumn{2}{c}{74/69}
        & \multicolumn{1}{|l}{Single} & \multicolumn{2}{c}{56/67} & \multicolumn{2}{c}{70/56} & \multicolumn{2}{c}{79/77} & \multicolumn{2}{c}{63/75} & \multicolumn{2}{c}{69/68} \\ \cmidrule(lr){2-12} \cmidrule(lr){13-23}

        & Mixtral 67/70 & \underline{67}/\textbf{74} & \textbf{72}/\textbf{74} & 70/\textbf{74} & 71/57 & 70/74 & 72/74 & 68/\underline{75} & 77/\underline{75} & 68/68 & \textbf{76}/69
        & 56/69 & 49/64 & 49/64 & 59/{\bf 72} & 54/68 & 59/72 & 54/68 & 43/74 & 49/\underline{75} & 46/65 & 56/67 \\

        & Mistral 65/71 & - & - & 69/56 & 71/56 & 70/74 & 71/74 & 69/\underline{75} & 75/\underline{75} & 67/68 & \textbf{75}/68
        & 56/67 & - & - & 54/{\bf 68} & 54/{\bf 68} & 54/68 & 54/68 & 49/\underline{75} & 49/\underline{75} & 56/67 & 56/67 \\

        & Gemma 73/57 & - & - & - & - & 70/74 & 71/74 & 73/\textbf{76} & \underline{78}/\textbf{78} & 72/\textbf{70} & \textbf{76}/\textbf{70}
        & 70/56 & - & - & - & - & 54/68 & 54/68 & 50/70 & 50/70 & 69/\underline{68} & 69/\underline{68} \\

        & Gemma-2 79/77 & - & - & - & - & - & - & 78/\textbf{78} & 78/\textbf{78} & 76/\textbf{78} & 74/\underline{77}
        & 79/77 & - & - & - & - & - & - & 50/70 & 50/70 & 43/75 & 43/75 \\

        & Llama-3 78/75 & - & - & - & - & - & - & - & - & 74/69 & 74/68
        & 63/75 & - & - & - & - & - & - & - & - & 56/67 & 56/67 \\

        \midrule

        \multirow{7}{*}{Metaculus}
        & Single & \multicolumn{2}{c}{61/70} & \multicolumn{2}{c}{76/73} & \multicolumn{2}{c}{68/66} & \multicolumn{2}{c}{81/65} & \multicolumn{2}{c}{81/74}
        & \multicolumn{1}{|l}{Single} & \multicolumn{2}{c}{61/68} & \multicolumn{2}{c}{76/73} & \multicolumn{2}{c}{68/66} & \multicolumn{2}{c}{81/65} & \multicolumn{2}{c}{82/76} \\ \cmidrule(lr){2-12} \cmidrule(lr){13-23}

        & Mixtral 66/74 & 63/\textbf{75} & 62/\textbf{75} & 72/73 & 74/\underline{74} & 65/66 & 66/66 & 66/\underline{74} & 66/\underline{74} & 73/\textbf{76} & 77/73
        & 61/73 & 57/{\bf 79} & 55/{\bf 78} & 63/72 & 68/{\bf 74} & 59/60 & 61/62 & 66/{\bf 76} & 66/{\bf 76} & 56/72 & 71/73 \\

        & Mistral 61/70 & - & - & 61/60 & 71/72 & 61/60 & 60/59 & 61/\underline{70} & 61/\underline{70} & 72/\underline{74} & 69/73
        & 62/67 & - & - & 65/{\bf 74} & 55/55 & 55/55 & 55/55 & 61/{\bf 74} & 61/{\bf 74} & 63/75 & 46/70 \\

        & Gemma 76/73 & - & - & - & - & 61/60 & 71/69 & 68/66 & 76/\underline{73} & 73/72 & 70/69
        & 71/73 & - & - & - & - & 55/55 & 61/61 & 76/\underline{73} & 68/66 & 62/74 & 56/56 \\

        & Gemma-2 68/66 & - & - & - & - & - & - & 68/\underline{66} & 68/\underline{66} & 73/72 & 70/69
        & 67/66 & - & - & - & - & - & - & 68/\underline{66} & 68/\underline{66} & 62/63 & 56/56 \\

        & Llama-3 81/65 & - & - & - & - & - & - & - & - & \underline{81}/\underline{74} & \underline{81}/\underline{74}
        & 78/65 & - & - & - & - & - & - & - & - & \underline{82}/\underline{76} & \underline{82}/\underline{76}  \\
    \end{tabular}
    \caption{Accuracy results of our multi-agent combination framework applied to pairs of ArgLLM agents for Depth 1 (left) and Depth 2 (right). We show accuracy results using  0.5/estimated base scores and $\omega_{avg}$ (Avg) and $\omega_{max}$ (Max) base score aggregations.
    Bold indicates improvement over both Single agents; underline indicates improvement over one and parity with the other.}
    \label{tab:combined}
\end{table*}

\subsection{RAG-ArgLLM Agent Pairs}
\label{app:ragargllm_pairs}
Table~\ref{tab:combined_nytimes} and Table~\ref{tab:combined_guardian} report results for the multi-agent framework using RAG-ArgLLM agents with NYTimes and Guardian (respectively) for the GJOpen and Metaculus datasets.

\begin{table*}[htp!]
    \centering
    \footnotesize
    \setlength{\tabcolsep}{0.28em}
    \begin{tabular}{crcccccccccc|crcccccccccc}
        & & \multicolumn{2}{c}{Mistral} & \multicolumn{2}{c}{Gemma} & \multicolumn{2}{c}{Gemma-2} & \multicolumn{2}{c}{Llama-3} & \multicolumn{2}{c|}{GPT-4o}
        & & \multicolumn{2}{c}{Mistral} & \multicolumn{2}{c}{Gemma} & \multicolumn{2}{c}{Gemma-2} & \multicolumn{2}{c}{Llama-3} & \multicolumn{2}{c}{GPT-4o} \\
        & & Avg & Max & Avg & Max & Avg & Max & Avg & Max & Avg & Max
        & & Avg & Max & Avg & Max & Avg & Max & Avg & Max & Avg & Max \\
        \midrule

        \multirow{7}{*}{GJOpen}
        & Single 
        & \multicolumn{2}{c}{59/68}
        & \multicolumn{2}{c}{71/66}
        & \multicolumn{2}{c}{76/75}
        & \multicolumn{2}{c}{78/44}
        & \multicolumn{2}{c}{72/66} &
        \multicolumn{1}{|l}{Single}
        & \multicolumn{2}{c}{52/68}
        & \multicolumn{2}{c}{64/63}
        & \multicolumn{2}{c}{75/75}
        & \multicolumn{2}{c}{75/50}
        & \multicolumn{2}{c}{71/66} \\ \cmidrule(lr){2-12} \cmidrule(lr){13-23}

        & Mixtral 61/71 
        & {\bf 62}/{\bf 74} & {\bf 67}/{\bf 73} 
        & 63/65 & 63/70 
        & 61/70 & 63/70 
        & 61/{\bf 71} & 61/\underline{71} 
        & 64/66 & \underline{72}/67 &
        49/67
        & 40/63 & 50/67 
        & 48/59 & 43/54 
        & 48/59 & 43/54 
        & 37/49 & 48/51 
        & 49/59 & 49/59 \\

        & Mistral 59/68 
        & - & - 
        & 66/{\bf 72} & 68/66 
        & 66/72 & 68/72 
        & 65/{\bf 73} & 65/{\bf 73} 
        & 66/67 & \underline{72}/66 &
        52/68
        & - & - 
        & 48/56 & 46/59 
        & 49/62 & 46/59 
        & 41/49 & 48/50 
        & 52/62 & 52/62 \\

        & Gemma 71/66 
        & - & - 
        & - & - 
        & 66/72 & 63/70 
        & 76/{\bf 75} & 76/{\bf 75} 
        & 71/{\bf 73} & 71/{\bf 68} &
        64/63 
        & - & - 
        & - & - 
        & \underline{75}/\underline{75} & 46/59
        & 53/54 & 37/57 
        & 64/{\bf 67} & 64/{\bf 67} \\

        & Gemma-2 76/75 
        & - & - 
        & - & - 
        & - & - 
        & 76/\underline{75} & 37/57
        & 71/73 & \underline{75}/\underline{75} &
        75/75 
        & - & - 
        & - & - 
        & - & - 
        & 66/71 & 76/\underline{75} 
        & \underline{75}/\underline{75} & 37/56 \\

        & Llama-3 78/44 
        & - & - 
        & - & - 
        & - & - 
        & - & - 
        & 72/\underline{66} & 72/\underline{66} &
        75/50
        & - & - 
        & - & - 
        & - & - 
        & - & - 
        & \underline{75}/{\bf 68} & \underline{75}/{\bf 68} \\

        \midrule

        \multirow{7}{*}{Metaculus}
        & Single 
        & \multicolumn{2}{c}{79/80}
        & \multicolumn{2}{c}{81/73}
        & \multicolumn{2}{c}{81/78}
        & \multicolumn{2}{c}{79/68}
        & \multicolumn{2}{c}{78/73} &
        \multicolumn{1}{|l}{Single} 
        & \multicolumn{2}{c}{61/78}
        & \multicolumn{2}{c}{72/74}
        & \multicolumn{2}{c}{76/74}
        & \multicolumn{2}{c}{78/66}
        & \multicolumn{2}{c}{79/73} \\ \cmidrule(lr){2-12} \cmidrule(lr){13-23}

        & Mixtral 71/78 
        & 76/{\bf 81} & 76/{\bf 81} 
        & 71/76 & 71/76 
        & 71/72 & 71/72 
        & 71/\underline{78} & 71/\underline{78} 
        & 74/76 & 75/77 &
        52/72 
        & 48/70 & 56/72
        & 49/69 & 53/68
        & 52/55 & 47/51
        & 41/63 & 52/66
        & 52/66 & 52/66 \\

        & Mistral 79/80 
        & - & - 
        & 79/77 & 79/76 
        & 79/77 & 79/77 
        & \underline{79}/\underline{80} & \underline{79}/\underline{80} 
        & 78/77 & 77/78 &
        61/78
        & - & - 
        & 62/72 & 61/64 
        & 61/63 & 61/64 
        & 53/67 & 56/65
        & 61/69 & 61/69 \\

        & Gemma 81/73 
        & - & - 
        & - & - 
        & 79/77 & \underline{81}/\underline{78} 
        & \underline{81}/78 & \underline{81}/\underline{73} 
        & 78/{\bf 77} & 78/{\bf 77} &
        72/74 
        & - & - 
        & - & - 
        & 61/63 & 69/68
        & 64/70 & 44/47
        & 72/\underline{74} & 76/\underline{74} \\

        & Gemma-2 81/78 
        & - & - 
        & - & - 
        & - & - 
        & \underline{81}/78 & \underline{81}/\underline{78} 
        & 78/77 & 78/77 &
        76/74 
        & - & - 
        & - & - 
        & - & - 
        & 71/70 & 44/47
        & 76/\underline{74} & 76/\underline{74} \\

        & Llama-3 79/68 
        & - & - 
        & - & - 
        & - & - 
        & - & - 
        & 78/\underline{73} & 78/\underline{73} &
        78/66
        & - & - 
        & - & - 
        & - & - 
        & - & - 
        & 78/72 & 78/72

    \end{tabular}
    Accuracy results of our multi-agent combination framework applied to pairs of ArgLLM agents for Depth 1 (left) and Depth 2 (right). We show accuracy results using 0.5/estimated base scores and $\omega_{avg}$ (Avg) and $\omega_{max}$ (Max) base score aggregations.
    Bold indicates improvement over both Single agents; underline indicates improvement over one and parity with the other.
    \caption{Accuracy results of our multi-agent combination framework applied to pairs of RAG-ArgLLM agents with NYTimes for Depth 1 (left) and Depth 2 (right). We show accuracy results using 0.5/estimated base scores and $\omega_{avg}$ (Avg) and $\omega_{max}$ (Max) base score aggregations. Bold indicates improvement over both Single agents; underline indicates improvement over one and parity with the other.}
    \label{tab:combined_nytimes}
\end{table*}

\begin{table*}[htp!]
    \centering
    \footnotesize
    \setlength{\tabcolsep}{0.28em}
    \begin{tabular}{crcccccccccc|crcccccccccc}
        & & \multicolumn{2}{c}{Mistral} & \multicolumn{2}{c}{Gemma} & \multicolumn{2}{c}{Gemma-2} & \multicolumn{2}{c}{Llama-3} & \multicolumn{2}{c|}{GPT-4o}
        & & \multicolumn{2}{c}{Mistral} & \multicolumn{2}{c}{Gemma} & \multicolumn{2}{c}{Gemma-2} & \multicolumn{2}{c}{Llama-3} & \multicolumn{2}{c}{GPT-4o} \\
        & & Avg & Max & Avg & Max & Avg & Max & Avg & Max & Avg & Max
        & & Avg & Max & Avg & Max & Avg & Max & Avg & Max & Avg & Max \\
        \midrule

        \multirow{7}{*}{GJOpen}
        & Single 
        & \multicolumn{2}{c}{66/74}
        & \multicolumn{2}{c}{72/64}
        & \multicolumn{2}{c}{78/76}
        & \multicolumn{2}{c}{78/76}
        & \multicolumn{2}{c}{71/67} &
        \multicolumn{1}{|l}{Single} 
        & \multicolumn{2}{c}{51/68}
        & \multicolumn{2}{c}{64/63}
        & \multicolumn{2}{c}{77/76}
        & \multicolumn{2}{c}{74/52}
        & \multicolumn{2}{c}{72/67} \\ \cmidrule(lr){2-12} \cmidrule(lr){13-23}

        & Mixtral 61/72 
        & 63/\underline{74} & {\bf 67}/\underline{74} 
        & 63/63 & 64/70 
        & 63/70 & 64/70 
        & 61/72 & 61/72 
        & 64/67 & \underline{71}/67 &
        48/68
        & 38/62 & 47/67 
        & 48/60 & 41/55
        & 48/60 & 41/55
        & 36/50 & 51/52
        & 49/60 & 49/60 \\

        & Mistral 66/74 
        & - & - 
        & 68/72 & 70/64 
        & 68/72 & 69/72 
        & 66/74 & 66/74 
        & 66/67 & \underline{71}/67 &
        51/68 
        & - & - 
        & 48/57 & 46/57 
        & 51/62 & 46/57 
        & 39/51 & 48/51
        & 51/61 & 56/66 \\

        & Gemma 72/64 
        & - & - 
        & - & - 
        & 68/72 & 64/70 
        & \underline{78}/\underline{76} & \underline{78}/\underline{76} 
        & 71/{\bf 74} & \underline{72}/{\bf 69} &
        64/63 
        & - & - 
        & - & - 
        & \underline{77}/\underline{76} & 46/57
        & 52/56 & 38/59
        & 64/\underline{67} & \underline{77}/\underline{76} \\

        & Gemma-2 78/76 
        & - & - 
        & - & - 
        & - & - 
        & \underline{78}/\underline{76} & \underline{78}/\underline{76} 
        & 71/74 & 72/74 &
        77/76 
        & - & - 
        & - & - 
        & - & - 
        & 71/73 & 38/59
        & \underline{77}/\underline{76} & \underline{77}/\underline{76} \\

        & Llama-3 78/76 
        & - & - 
        & - & - 
        & - & - 
        & - & - 
        & 71/67 & 71/67 &
        74/52 
        & - & - 
        & - & - 
        & - & - 
        & - & - 
        & \underline{74}/{\bf 68} & \underline{74}/{\bf 68} \\

        \midrule

        \multirow{7}{*}{Metaculus}
        & Single 
        & \multicolumn{2}{c}{79/80}
        & \multicolumn{2}{c}{80/76}
        & \multicolumn{2}{c}{81/78}
        & \multicolumn{2}{c}{79/70}
        & \multicolumn{2}{c}{79/76} &
        \multicolumn{1}{|l}{Single}
        & \multicolumn{2}{c}{66/77}
        & \multicolumn{2}{c}{68/73}
        & \multicolumn{2}{c}{80/78}
        & \multicolumn{2}{c}{73/68}
        & \multicolumn{2}{c}{80/76} \\ \cmidrule(lr){2-12} \cmidrule(lr){13-23}

        & Mixtral 69/73
        & 72/79 & 73/79 
        & 71/74 & 70/73 
        & 68/68 & 69/68 
        & 69/\underline{73} & 69/\underline{73} 
        & 71/75 & 73/\underline{76} &
        55/73
        & 53/74 & 60/74 
        & 54/68 & 53/68 
        & 54/56 & 44/48 
        & 45/64 & 55/66
        & 55/69 & 55/69 \\

        & Mistral 79/80 
        & - & - 
        & 79/77 & 79/76 
        & 79/77 & 79/77 
        & \underline{79}/\underline{80} & \underline{79}/\underline{80} 
        & \underline{79}/77 & {\bf 80}/77 &
        66/77 
        & - & - 
        & 63/71 & 63/63 
        & 66/66 & 63/63 
        & 56/66 & 61/66
        & 66/71 & 66/71 \\

        & Gemma 80/76 
        & - & - 
        & - & - 
        & 79/77 & 80/\underline{78} 
        & {\bf 81}/{\bf 78} & \underline{80}/\underline{76} 
        & 79/{\bf 78} & 79/{\bf 78} &
        68/73 
        & - & - 
        & - & - 
        & 66/66 & 64/63 
        & 62/69 & 45/48
        & 68/72 & \underline{80}/{\bf 78} \\

        & Gemma-2 81/78 
        & - & - 
        & - & - 
        & - & - 
        & \underline{81}/\underline{78} & \underline{81}/\underline{78} 
        & 79/\underline{78} & 79/\underline{78} &
        80/78 
        & - & - 
        & - & - 
        & - & - 
        & 76/75 & 45/48 
        & \underline{80}/\underline{78} & \underline{80}/\underline{78} \\

        & Llama-3 79/70 
        & - & - 
        & - & - 
        & - & - 
        & - & - 
        & \underline{79}/\underline{76} & \underline{79}/\underline{76} &
        73/68 
        & - & - 
        & - & - 
        & - & - 
        & - & - 
        & 73/\underline{76} & 73/\underline{76}

    \end{tabular}
    \caption{Accuracy results of our multi-agent combination framework applied to pairs of RAG-ArgLLM agents with Guardian for Depth 1 (left) and Depth 2 (right). We show accuracy results using 0.5/estimated base scores and $\omega_{avg}$ (Avg) and $\omega_{max}$ (Max) base score aggregations. Bold indicates improvement over both Single agents; underline indicates improvement over one and parity with the other.}
    \label{tab:combined_guardian}
\end{table*}

\subsection{Two-Agent Combination}
\label{app:two-agent}
Tables~\ref{tab:combined_nytimes_guardian_d1} and \ref{tab:combined_nytimes_guardian_d2} show the results of combining two RAG-ArgLLM agents, where one uses NYTimes and the other uses Guardian as external sources.
Tables~\ref{tab:combined_nytimes_argllm_d1} and \ref{tab:combined_nytimes_argllm_d2} present results where ArgLLM agents are paired with RAG-ArgLLM agents using NYTimes. Tables~\ref{tab:combined_guardian_argllm_d1} and \ref{tab:combined_guardian_argllm_d2} shows similar pairings but with Guardian as the external source.

\begin{table*}[bt]
    \centering
\begin{tabular}{crcccccccccccc}
        & & \multicolumn{2}{c}{Mixtral} 
          & \multicolumn{2}{c}{Mistral} 
          & \multicolumn{2}{c}{Gemma} 
          & \multicolumn{2}{c}{Gemma-2} 
          & \multicolumn{2}{c}{Llama-3} 
          & \multicolumn{2}{c}{GPT-4o} \\
        \cmidrule(lr){3-4} 
        \cmidrule(lr){5-6} 
        \cmidrule(lr){7-8} 
        \cmidrule(lr){9-10} 
        \cmidrule(lr){11-12} 
        \cmidrule(lr){13-14}
        & & Avg & Max & Avg & Max & Avg & Max & Avg & Max & Avg & Max & Avg & Max \\
        \midrule

        \multirow{7}{*}{GJOpen} 
            & Single     
                & 61/72 & 61/72 & 66/74 & 66/74 & 72/64 & 72/64 & 78/76 & 78/76 & 78/76 & 78/76 & 71/67 & 71/67 
\\ \cmidrule{2-14}

            & Mixtral 61/71    
                & \underline{61}/\underline{72} & \underline{61}/\underline{72} & 63/\underline{74} & {\bf 68}/\underline{74} & 64/63 & 63/69 & 62/69 & 63/69 & 62/69 & 63/69 & 63/67 & \underline{71}/67 
\\

            & Mistral 59/68    
                & {\bf 62}/{\bf 74} & {\bf 67}/{\bf 73} & \underline{66}/\underline{74} & \underline{66}/\underline{74} & 66/{\bf 71} & 69/64 & 66/71 & 67/72 & 66/71 & 67/72 & 66/\underline{68} & {\bf 72}/67 
\\

            & Gemma 71/66       
                & 63/{\bf 74} & 63/{\bf 73} & 67/\underline{74} & 68/\underline{74} & {\bf 78}/{\bf 76} & {\bf 78}/{\bf 76} & \underline{78}/\underline{76} & \underline{78}/\underline{76} & 72/74 & 72/74 & \underline{71}/{\bf 68} & {\bf 72}/{\bf 68} 
\\

            & Gemma-2 76/75     
                & 61/73 & 63/73 & 67/74 & 68/74 & {\bf 78}/{\bf 76} & {\bf 78}/{\bf 76} & \underline{78}/\underline{76} & \underline{78}/\underline{76} & 76/75 & 76/75 & 71/68 & 71/68 
\\

            & Llama-3 78/44     
                & 61/{\bf 75} & 76/{\bf 75} & 66/{\bf 76} & 76/{\bf 77} & 76/{\bf 75} & 76/{\bf 75} & 76/75 & 76/75 & \underline{78}/\underline{76} & \underline{78}/\underline{76} & 72/\underline{67} & 72/\underline{67} 
\\
\end{tabular}
    \caption{Accuracy results of our multi-agent combination framework applied to RAG-ArgLLM agents with Guardian (columns) and RAG-ArgLLM agents with NYTimes (rows) for Depth 1. We show accuracy results using 0.5/estimated base scores and $\omega_{avg}$ (Avg) and $\omega_{max}$ (Max) base score aggregations. Bold indicates improvement over both Single agents; underline indicates improvement over one and parity with the other.}
    \label{tab:combined_nytimes_guardian_d1}
\end{table*}

\begin{table*}[bt]
    \centering
    \begin{tabular}{crcccccccccccc}
        & & \multicolumn{2}{c}{Mixtral} 
          & \multicolumn{2}{c}{Mistral} 
          & \multicolumn{2}{c}{Gemma} 
          & \multicolumn{2}{c}{Gemma-2} 
          & \multicolumn{2}{c}{Llama-3} 
          & \multicolumn{2}{c}{GPT-4o} \\
        \cmidrule(lr){3-4} 
        \cmidrule(lr){5-6} 
        \cmidrule(lr){7-8} 
        \cmidrule(lr){9-10} 
        \cmidrule(lr){11-12} 
        \cmidrule(lr){13-14}
        & & Avg & Max & Avg & Max & Avg & Max & Avg & Max & Avg & Max & Avg & Max \\
        \midrule

        \multirow{7}{*}{GJOpen} 
            & Single & 49/68 & 49/68 & 51/68 & 51/68 & 64/63 & 64/63 & 77/76 & 77/76 & 74/52 & 74/52 & 72/67 & 72/67 \\ \cmidrule{2-14}

            & Mixtral 49/67 & 34/57 & 42/60 & 40/62 & 48/66 & 49/59 & 42/54 & 49/59 & 42/54 & 38/50 & 49/52 & 36/52 & 48/57 \\

            & Mistral 52/68 & 37/63 & 48/66 & 40/65 & 50/\underline{68} & 49/59 & 46/59 & 52/63 & 46/59 & 41/50 & 49/51 & 39/60 & 50/63 \\

            & Gemma 64/63 & 49/{\bf 69} & 46/{\bf 70} & 46/{\bf 69} & 50/{\bf 69} & {\bf 73}/{\bf 74} & {\bf 69}/{\bf 70} & 73/74 & 69/70 & 53/54 & 37/50 & 65/{\bf 69} & 36/60 \\

            & Gemma-2 75/75 & 49/69 & 41/65 & 51/68 & 46/63 & 73/74 & 69/70 & 73/74 & 69/70 & 67/55 & 37/50 & 65/69 & 36/60 \\

            & Llama-3 75/50 & 32/{\bf 70} & 36/{\bf 74} & 37/{\bf 70} & 39/{\bf 73} & 71/{\bf 73} & 36/56 & 71/73 & 36/56 & 34/50 & 28/49 & 35/64 & 37/64 \\
\end{tabular}
    \caption{Accuracy results of our multi-agent combination framework applied to RAG-ArgLLM agents with Guardian (columns) and RAG-ArgLLM agents with NYTimes (rows) for Depth 2. We show accuracy results using 0.5/estimated base scores and $\omega_{avg}$ (Avg) and $\omega_{max}$ (Max) base score aggregations. Bold indicates improvement over both Single agents; underline indicates improvement over one and parity with the other.}
    \label{tab:combined_nytimes_guardian_d2}
\end{table*}

\begin{table*}[bt]
    \centering
    \begin{tabular}{crcccccccccccc}
        & & \multicolumn{2}{c}{Mixtral} 
          & \multicolumn{2}{c}{Mistral} 
          & \multicolumn{2}{c}{Gemma} 
          & \multicolumn{2}{c}{Gemma-2} 
          & \multicolumn{2}{c}{Llama-3} 
          & \multicolumn{2}{c}{GPT-4o} \\
        \cmidrule(lr){3-4} \cmidrule(lr){5-6} \cmidrule(lr){7-8} 
        \cmidrule(lr){9-10} \cmidrule(lr){11-12} \cmidrule(lr){13-14}
        & & Avg & Max & Avg & Max & Avg & Max & Avg & Max & Avg & Max & Avg & Max \\
        \midrule

        \multirow{7}{*}{GJOpen} 
        & Single     
            & 67/70 & 67/70 & 65/71 & 65/71 & 73/57 & 73/57 & 79/77 & 79/77 & 78/75 & 78/75 & 74/69 & 74/69 
\\ \cmidrule{2-14}

        & Mixtral 61/71    
            & \underline{67}/70 & \underline{67}/70 & 63/{\bf 72} & {\bf 67}/{\bf 72} & 65/56 & 69/{\bf 75} & 69/75 & 69/75 & 67/\underline{75} & 77/{\bf 76} & 65/68 & \underline{74}/68 
\\

        & Mistral 59/68    
            & 66/{\bf 73} & {\bf 72}/{\bf 73} & \underline{65}/\underline{71} & \underline{65}/\underline{71} & 70/{\bf 76} & 69/57 & 70/76 & 70/75 & 69/\underline{75} & 76/\underline{75} & 68/\underline{69} & \underline{74}/\underline{69} 
\\

        & Gemma 71/66       
            & 67/{\bf 73} & 68/{\bf 71} & 66/\underline{71} & 67/{\bf 72} & {\bf 79}/{\bf 77} & {\bf 79}/{\bf 77} & \underline{79}/\underline{77} & \underline{79}/\underline{77} & 72/{\bf 76} & 72/{\bf 76} & 73/\underline{69} & \underline{74}/{\bf 70} 
\\

        & Gemma-2 76/75     
            & 67/72 & 68/71 & 66/71 & 67/72 & {\bf 79}/{\bf 77} & {\bf 79}/{\bf 77} & \underline{79}/\underline{77} & \underline{79}/\underline{77} & 75/{\bf 76} & 75/{\bf 76} & 73/69 & 73/69 
\\

        & Llama-3 78/44     
            & 66/{\bf 75} & 77/{\bf 75} & 65/{\bf 75} & 74/{\bf 76} & \underline{78}/{\bf 78} & \underline{78}/{\bf 77} & 78/{\bf 78} & 78/\underline{77} & \underline{78}/\underline{75} & \underline{78}/\underline{75} & 76/\underline{69} & 74/68 
\\
\end{tabular}
    \caption{Accuracy results of our multi-agent combination framework applied to ArgLLM agents (columns) and RAG-ArgLLM agents with NYTimes (rows) for Depth 1. We show accuracy results using 0.5/estimated base scores and $\omega_{avg}$ (Avg) and $\omega_{max}$ (Max) base score aggregations. Bold indicates improvement over both Single agents; underline indicates improvement over one and parity with the other.}
    \label{tab:combined_nytimes_argllm_d1}
\end{table*}

\begin{table*}[bt]
    \centering
    \begin{tabular}{crcccccccccccc}
        & & \multicolumn{2}{c}{Mixtral} 
          & \multicolumn{2}{c}{Mistral} 
          & \multicolumn{2}{c}{Gemma} 
          & \multicolumn{2}{c}{Gemma-2} 
          & \multicolumn{2}{c}{Llama-3} 
          & \multicolumn{2}{c}{GPT-4o} \\
        \cmidrule(lr){3-4} \cmidrule(lr){5-6} \cmidrule(lr){7-8} 
        \cmidrule(lr){9-10} \cmidrule(lr){11-12} \cmidrule(lr){13-14}
        & & Avg & Max & Avg & Max & Avg & Max & Avg & Max & Avg & Max & Avg & Max \\
        \midrule

        \multirow{7}{*}{GJOpen} 
        & Single     
            & 56/69 & 56/69 & 56/67 & 56/67 & 70/56 & 70/56 & 79/77 & 79/77 & 63/75 & 63/75 & 69/68 & 69/68 \\ \cmidrule{2-14}

        & Mixtral 61/71    
            & 41/66 & 54/67 & 44/64 & 56/{\bf 69} & 55/{\bf 69} & 52/\underline{67} & 55/69 & 52/67 & 37/70 & 46/74 & 38/62 & 63/\underline{68} \\

        & Mistral 59/68    
            & 42/68 & 55/{\bf 70} & 45/65 & 55/{\bf 69} & 55/52 & 53/67 & 56/71 & 53/67 & 42/73 & 47/\underline{75} & 42/65 & 59/67 \\

        & Gemma 71/66       
            & 56/{\bf 70} & 49/{\bf 71} & 52/{\bf 70} & 53/{\bf 70} & {\bf 75}/{\bf 76} & \underline{70}/{\bf 75} & 75/76 & 70/75 & 52/74 & 42/73 & 68/{\bf 70} & 33/63 \\

        & Gemma-2 76/75     
            & 56/70 & 45/67 & 56/69 & 50/64 & \underline{75}/{\bf 76} & 70/\underline{75} & 75/76 & 70/75 & 65/{\bf 76} & 42/73 & 68/70 & 33/63 \\

        & Llama-3 78/44     
            & 41/{\bf 72} & 45/{\bf 74} & 46/{\bf 71} & 48/{\bf 73} & 72/{\bf 77} & 48/{\bf 75} & 72/\underline{77} & 48/75 & 31/\underline{75} & 31/\underline{75} & 37/66 & 46/66 \\
\end{tabular}
    \caption{Accuracy results of our multi-agent combination framework applied to ArgLLM agents (columns) and RAG-ArgLLM agents with NYTimes (rows) for Depth 2. We show accuracy results using 0.5/estimated base scores and $\omega_{avg}$ (Avg) and $\omega_{max}$ (Max) base score aggregations. Bold indicates improvement over both Single agents; underline indicates improvement over one and parity with the other.}
    \label{tab:combined_nytimes_argllm_d2}
\end{table*}

\begin{table*}[bt]
    \centering
    \begin{tabular}{crcccccccccccc}
        & & \multicolumn{2}{c}{Mixtral} 
          & \multicolumn{2}{c}{Mistral} 
          & \multicolumn{2}{c}{Gemma} 
          & \multicolumn{2}{c}{Gemma-2} 
          & \multicolumn{2}{c}{Llama-3} 
          & \multicolumn{2}{c}{GPT-4o} \\
        \cmidrule(lr){3-4} \cmidrule(lr){5-6} \cmidrule(lr){7-8} 
        \cmidrule(lr){9-10} \cmidrule(lr){11-12} \cmidrule(lr){13-14}
        & & Avg & Max & Avg & Max & Avg & Max & Avg & Max & Avg & Max & Avg & Max \\
        \midrule

        \multirow{7}{*}{GJOpen} 
        & Single     
            & 67/70 & 67/70 & 65/71 & 65/71 & 73/57 & 73/57 & 79/77 & 79/77 & 78/75 & 78/75 & 74/69 & 74/69 
\\ \cmidrule{2-14}

        & Mixtral 61/72    
            & \underline{67}/70 & \underline{67}/70 & 62/\underline{72} & {\bf 67}/\underline{72} & 65/56 & 69/{\bf 75} & 67/76 & 69/75 & 67/\underline{75} & 77/\underline{75} & 66/68 & {\bf 75}/69 
\\

        & Mistral 66/74    
            & 66/\underline{74} & {\bf 71}/73 & 65/71 & 65/71 & 71/{\bf 76} & 69/57 & 71/76 & 71/76 & 70/{\bf 76} & 77/\underline{75} & 69/68 & \underline{74}/68 
\\

        & Gemma 72/64       
            & 68/{\bf 74} & 69/{\bf 71} & 67/\underline{71} & 67/\underline{71} & {\bf 79}/{\bf 77} & {\bf 79}/{\bf 77} & \underline{79}/\underline{77} & \underline{79}/\underline{77} & 72/{\bf 76} & 72/{\bf 76} & \underline{74}/{\bf 70} & \underline{74}/{\bf 70} 
\\

        & Gemma-2 78/76     
            & 68/71 & 69/71 & 67/71 & 67/71 & {\bf 79}/{\bf 77} & {\bf 79}/{\bf 77} & \underline{79}/\underline{77} & \underline{79}/\underline{77} & 77/\underline{76} & 77/\underline{76} & 74/70 & 74/70 
\\

        & Llama-3 78/76     
            & 68/71 & 69/71 & 67/71 & 67/71 & \underline{78}/{\bf 77} & \underline{78}/{\bf 77} & 78/\underline{77} & 78/\underline{77} & \underline{78}/75 & \underline{78}/75 & 74/70 & 74/70 
\\
\end{tabular}
    \caption{Accuracy results of our multi-agent combination framework applied to ArgLLM agents (columns) and RAG-ArgLLM agents with Guardian (rows) for Depth 1. We show accuracy results using 0.5/estimated base scores and $\omega_{avg}$ (Avg) and $\omega_{max}$ (Max) base score aggregations. Bold indicates improvement over both Single agents; underline indicates improvement over one and parity with the other.}
    \label{tab:combined_guardian_argllm_d1}
\end{table*}

\begin{table*}[bt]
    \centering
    \begin{tabular}{crcccccccccccc}
        & & \multicolumn{2}{c}{Mixtral} 
          & \multicolumn{2}{c}{Mistral} 
          & \multicolumn{2}{c}{Gemma} 
          & \multicolumn{2}{c}{Gemma-2} 
          & \multicolumn{2}{c}{Llama-3} 
          & \multicolumn{2}{c}{GPT-4o} \\
        \cmidrule(lr){3-4} \cmidrule(lr){5-6} \cmidrule(lr){7-8} 
        \cmidrule(lr){9-10} \cmidrule(lr){11-12} \cmidrule(lr){13-14}
        & & Avg & Max & Avg & Max & Avg & Max & Avg & Max & Avg & Max & Avg & Max \\
        \midrule

        \multirow{7}{*}{GJOpen} 
        & Single     
            & 56/69 & 56/69 & 56/67 & 56/67 & 70/56 & 70/56 & 79/77 & 79/77 & 63/75 & 63/75 & 69/68 & 69/68 
\\ \cmidrule{2-14}

        & Mixtral 61/72    
            & 39/65 & 54/67 & 42/65 & 55/{\bf 69} & 54/{\bf 71} & 52/67 & 54/71 & 52/67 & 35/72 & 46/\underline{75} & 37/63 & 65/\underline{68} 
\\

        & Mistral 66/74    
            & 42/68 & 54/{\bf 70} & 43/65 & 55/\underline{68} & 54/52 & 53/66 & 54/70 & 53/66 & 39/74 & 46/\underline{75} & 40/64 & 60/67 
\\

        & Gemma 72/64       
            & 57/{\bf 71} & 49/{\bf 71} & 52/{\bf 69} & 52/{\bf 69} & {\bf 77}/{\bf 77} & {\bf 74}/{\bf 76} & 77/\underline{77} & 74/76 & 51/\underline{75} & 42/73 & {\bf 73}/{\bf 71} & 35/64 
\\

        & Gemma-2 78/76     
            & 57/71 & 46/67 & 57/69 & 49/63 & \underline{77}/{\bf 77} & 74/\underline{76} & 77/\underline{77} & 74/76 & 68/\underline{76} & 42/73 & 73/71 & 35/64 
\\

        & Llama-3 78/76     
            & 43/{\bf 73} & 49/{\bf 75} & 47/{\bf 71} & 50/{\bf 73} & 73/{\bf 77} & 49/{\bf 74} & 73/\underline{77} & 49/74 & 32/\underline{75} & 33/\underline{75} & 38/66 & 50/66 
\\
\end{tabular}
    \caption{Accuracy results of our multi-agent combination framework applied to ArgLLM agents (columns) and RAG-ArgLLM agents with Guardian (rows) for Depth 2. We show accuracy results using 0.5/estimated base scores and $\omega_{avg}$ (Avg) and $\omega_{max}$ (Max) base score aggregations. Bold indicates improvement over both Single agents; underline indicates improvement over one and parity with the other.}
    \label{tab:combined_guardian_argllm_d2}
\end{table*}

\subsection{Three-Agent Combination}
\label{app:three-agent}
The results, presented in Tables~\ref{tab:d1_argllm_mixtral_nytimes_guardian}-\ref{tab:d2_argllm_gpt4o_nytimes_guardian}, demonstrate the three agent combinations.

\begin{table*}[bt]
    \centering
\begin{tabular}{crcccccccccccc}
        & & \multicolumn{2}{c}{Mixtral} 
          & \multicolumn{2}{c}{Mistral} 
          & \multicolumn{2}{c}{Gemma} 
          & \multicolumn{2}{c}{Gemma-2} 
          & \multicolumn{2}{c}{Llama-3} 
          & \multicolumn{2}{c}{GPT-4o} \\
        \cmidrule(lr){3-4} \cmidrule(lr){5-6} \cmidrule(lr){7-8} 
        \cmidrule(lr){9-10} \cmidrule(lr){11-12} \cmidrule(lr){13-14}
        & & Avg & Max & Avg & Max & Avg & Max & Avg & Max & Avg & Max & Avg & Max \\
        \midrule

        \multirow{7}{*}{GJOpen - 67/70} 
        & Single     
            & 61/71 & 61/71 & 59/68 & 59/68 & 71/66 & 71/66 & 76/75 & 76/75 & 78/44 & 78/44 & 72/66 & 72/66 
\\ \cmidrule{2-14}

        & Mixtral 61/72    
            & {\bf 68}/70 & {\bf 68}/70 & {\bf 68}/70 & {\bf 68}/70 & 68/70 & 68/70 & 68/70 & 68/70 & 68/70 & 68/70 & 68/70 & 68/70 
\\

        & Mistral 66/74    
            & {\bf 71}/73 & {\bf 71}/{\bf 73} & {\bf 71}/73 & {\bf 71}/73 & \underline{71}/73 & 71/{\bf 73} & 71/73 & 71/73 & 71/73 & 71/73 & 71/73 & \underline{71}/{\bf 73} 
\\

        & Gemma 72/64       
            & 69/\underline{71} & 69/71 & 69/{\bf 71} & 69/71 & 69/{\bf 74} & 69/{\bf 74} & 69/74 & 69/74 & 69/{\bf 74} & 69/74 & 69/{\bf 71} & 69/{\bf 71} 
\\

        & Gemma-2 78/76     
            & 69/71 & 69/71 & 69/71 & 69/71 & 69/71 & 69/71 & 69/71 & 69/71 & 69/71 & 69/71 & 69/71 & 69/71 
\\

        & Llama-3 78/76     
            & 69/71 & 69/71 & 69/71 & 69/71 & 69/71 & 69/{\bf 71} & 69/71 & 69/71 & 69/71 & 69/71 & 69/71 & 69/{\bf 71} 
\\
\end{tabular}
    \caption{Accuracy results of our multi-agent combination framework applied to ArgLLM with Mixtral, RAG-ArgLLM agents with NYTimes (columns), and RAG-ArgLLM agents with Guardian (rows) for Depth 1. We show accuracy results using 0.5/estimated base scores and $\omega_{avg}$ (Avg) and $\omega_{max}$ (Max) base score aggregations. Bold indicates improvement over both Single agents; underline indicates improvement over one and parity with the other.}
    \label{tab:d1_argllm_mixtral_nytimes_guardian}
\end{table*}

\begin{table*}[bt]
    \centering
    \begin{tabular}{crcccccccccccc}
        & & \multicolumn{2}{c}{Mixtral} 
          & \multicolumn{2}{c}{Mistral} 
          & \multicolumn{2}{c}{Gemma} 
          & \multicolumn{2}{c}{Gemma-2} 
          & \multicolumn{2}{c}{Llama-3} 
          & \multicolumn{2}{c}{GPT-4o} \\
        \cmidrule(lr){3-4} \cmidrule(lr){5-6} \cmidrule(lr){7-8} 
        \cmidrule(lr){9-10} \cmidrule(lr){11-12} \cmidrule(lr){13-14}
        & & Avg & Max & Avg & Max & Avg & Max & Avg & Max & Avg & Max & Avg & Max \\
        \midrule

        \multirow{7}{*}{GJOpen - 65/71} 
        & Single     
            & 61/71 & 61/71 & 59/68 & 59/68 & 71/66 & 71/66 & 76/75 & 76/75 & 78/44 & 78/44 & 72/66 & 72/66 
\\ \cmidrule{2-14}

        & Mixtral 61/72    
            & {\bf 67}/\underline{72} & {\bf 67}/\underline{72} & {\bf 67}/\underline{72} & {\bf 67}/72 & 67/\underline{72} & 67/{\bf 72} & 67/72 & 67/72 & 67/\underline{72} & 67/72 & 67/\underline{72} & 67/{\bf 72} 
\\

        & Mistral 66/74    
            & {\bf 70}/73 & {\bf 70}/{\bf 73} & {\bf 70}/73 & {\bf 70}/73 & 70/73 & 70/{\bf 73} & 70/73 & 70/73 & 70/73 & 70/73 & 70/73 & 70/{\bf 73} 
\\

        & Gemma 72/64       
            & 69/{\bf 74} & 69/{\bf 74} & 67/\underline{71} & 67/71 & 67/\underline{71} & 67/\underline{71} & 67/71 & 67/71 & 69/{\bf 74} & 69/74 & 69/{\bf 74} & 69/{\bf 74} 
\\

        & Gemma-2 78/76     
            & 67/71 & 67/71 & 67/71 & 67/71 & 67/71 & 67/71 & 67/71 & 67/71 & 67/71 & 67/71 & 67/71 & 67/71 
\\

        & Llama-3 78/76     
            & 67/71 & 67/71 & 67/71 & 67/71 & 67/71 & 67/\underline{71} & 67/71 & 67/71 & 67/71 & 67/71 & 67/71 & 67/\underline{71} 
\\
\end{tabular}
    \caption{Accuracy results of our multi-agent combination framework applied to ArgLLM with Mistral, RAG-ArgLLM agents with NYTimes (columns), and RAG-ArgLLM agents with Guardian (rows) for Depth 1. We show accuracy results using 0.5/estimated base scores and $\omega_{avg}$ (Avg) and $\omega_{max}$ (Max) base score aggregations. Bold indicates improvement over both Single agents; underline indicates improvement over one and parity with the other.}
    \label{tab:d1_argllm_mistral_nytimes_guardian}
\end{table*}

\begin{table*}[bt]
    \centering
    \begin{tabular}{crcccccccccccc}
        & & \multicolumn{2}{c}{Mixtral} 
          & \multicolumn{2}{c}{Mistral} 
          & \multicolumn{2}{c}{Gemma} 
          & \multicolumn{2}{c}{Gemma-2} 
          & \multicolumn{2}{c}{Llama-3} 
          & \multicolumn{2}{c}{GPT-4o} \\
        \cmidrule(lr){3-4} \cmidrule(lr){5-6} \cmidrule(lr){7-8} 
        \cmidrule(lr){9-10} \cmidrule(lr){11-12} \cmidrule(lr){13-14}
        & & Avg & Max & Avg & Max & Avg & Max & Avg & Max & Avg & Max & Avg & Max \\
        \midrule

        \multirow{7}{*}{GJOpen - 73/57} 
        & Single     
            & 61/71 & 61/71 & 59/68 & 59/68 & 71/66 & 71/66 & 76/75 & 76/75 & 78/44 & 78/44 & 72/66 & 72/66 
\\ \cmidrule{2-14}

        & Mixtral 61/72    
            & 69/{\bf 75} & 69/{\bf 75} & 66/56 & 66/56 & 69/{\bf 75} & 69/{\bf 75} & 66/56 & 66/56 & 69/{\bf 75} & 69/75 & 66/56 & 66/56 
\\

        & Mistral 66/74    
            & 69/57 & 69/57 & 71/{\bf 76} & 71/{\bf 76} & 71/{\bf 76} & 71/{\bf 76} & 69/57 & 69/57 & 69/57 & 69/57 & 69/57 & 69/57 
\\

        & Gemma 72/64       
            & \underline{73}/56 & \underline{73}/56 & \underline{73}/57 & \underline{73}/57 & {\bf 78}/{\bf 77} & {\bf 78}/{\bf 77} & 73/56 & 73/56 & 74/{\bf 77} & 74/{\bf 77} & \underline{73}/57 & \underline{73}/57 
\\

        & Gemma-2 78/76     
            & 73/57 & 73/57 & 73/57 & 73/57 & \underline{78}/{\bf 77} & {\bf 78}/{\bf 77} & 73/57 & 73/57 & 73/57 & 73/57 & 73/57 & 73/57 
\\

        & Llama-3 78/76     
            & 73/57 & 73/57 & \underline{78}/{\bf 77} & \underline{78}/{\bf 77} & \underline{78}/{\bf 77} & \underline{78}/{\bf 77} & 73/57 & 73/57 & \underline{78}/{\bf 77} & \underline{78}/{\bf 77} & 73/57 & 73/57 
\\
\end{tabular}
    \caption{Accuracy results of our multi-agent combination framework applied to ArgLLM with Gemma, RAG-ArgLLM agents with NYTimes (columns), and RAG-ArgLLM agents with Guardian (rows) for Depth 1. We show accuracy results using 0.5/estimated base scores and $\omega_{avg}$ (Avg) and $\omega_{max}$ (Max) base score aggregations. Bold indicates improvement over both Single agents; underline indicates improvement over one and parity with the other.}
    \label{tab:d1_argllm_gemma_nytimes_guardian}
\end{table*}

\begin{table*}[bt]
    \centering
    \begin{tabular}{crcccccccccccc}
        & & \multicolumn{2}{c}{Mixtral} 
          & \multicolumn{2}{c}{Mistral} 
          & \multicolumn{2}{c}{Gemma} 
          & \multicolumn{2}{c}{Gemma-2} 
          & \multicolumn{2}{c}{Llama-3} 
          & \multicolumn{2}{c}{GPT-4o} \\
        \cmidrule(lr){3-4} \cmidrule(lr){5-6} \cmidrule(lr){7-8} 
        \cmidrule(lr){9-10} \cmidrule(lr){11-12} \cmidrule(lr){13-14}
        & & Avg & Max & Avg & Max & Avg & Max & Avg & Max & Avg & Max & Avg & Max \\
        \midrule

        \multirow{7}{*}{GJOpen - 79/77} 
        & Single     
            & 61/71 & 61/71 & 59/68 & 59/68 & 71/66 & 71/66 & 76/75 & 76/75 & 78/44 & 78/44 & 72/66 & 72/66 
\\ \cmidrule{2-14}

        & Mixtral 61/72    
            & 69/75 & 69/75 & 69/75 & 69/75 & 69/75 & 69/75 & 69/75 & 69/75 & 69/75 & 69/75 & 69/75 & 69/75 
\\

        & Mistral 66/74    
            & 71/76 & 71/76 & 71/76 & 71/76 & 71/76 & 71/76 & 71/76 & 71/76 & 71/76 & 71/76 & 71/76 & 71/76 
\\

        & Gemma 72/64       
            & 74/\underline{77} & 74/\underline{77} & 78/\underline{77} & 78/\underline{77} & 78/\underline{77} & 78/\underline{77} & 74/\underline{77} & 74/\underline{77} & 74/\underline{77} & 74/\underline{77} & 78/\underline{77} & 78/\underline{77} 
\\

        & Gemma-2 78/76     
            & 78/\underline{77} & 78/\underline{77} & 78/\underline{77} & 78/\underline{77} & 78/\underline{77} & 78/\underline{77} & 78/\underline{77} & 78/\underline{77} & 78/\underline{77} & 78/\underline{77} & 78/\underline{77} & 78/\underline{77} 
\\

        & Llama-3 78/76     
            & 78/\underline{77} & 78/\underline{77} & 78/\underline{77} & 78/\underline{77} & 78/\underline{77} & 78/\underline{77} & 78/\underline{77} & 78/\underline{77} & 78/\underline{77} & 78/\underline{77} & 78/\underline{77} & 78/\underline{77} 
\\
\end{tabular}
    \caption{Accuracy results of our multi-agent combination framework applied to ArgLLM with Gemma-2, RAG-ArgLLM agents with NYTimes (columns), and RAG-ArgLLM agents with Guardian (rows) for Depth 1. We show accuracy results using 0.5/estimated base scores and $\omega_{avg}$ (Avg) and $\omega_{max}$ (Max) base score aggregations. Bold indicates improvement over both Single agents; underline indicates improvement over one and parity with the other.}
    \label{tab:d1_argllm_gemma2_nytimes_guardian}
\end{table*}

\begin{table*}[bt]
    \centering
    \begin{tabular}{crcccccccccccc}
        & & \multicolumn{2}{c}{Mixtral} 
          & \multicolumn{2}{c}{Mistral} 
          & \multicolumn{2}{c}{Gemma} 
          & \multicolumn{2}{c}{Gemma-2} 
          & \multicolumn{2}{c}{Llama-3} 
          & \multicolumn{2}{c}{GPT-4o} \\
        \cmidrule(lr){3-4} \cmidrule(lr){5-6} \cmidrule(lr){7-8} 
        \cmidrule(lr){9-10} \cmidrule(lr){11-12} \cmidrule(lr){13-14}
        & & Avg & Max & Avg & Max & Avg & Max & Avg & Max & Avg & Max & Avg & Max \\
        \midrule

        \multirow{7}{*}{GJOpen - 78/75} 
        & Single     
            & 61/71 & 61/71 & 59/68 & 59/68 & 71/66 & 71/66 & 76/75 & 76/75 & 78/44 & 78/44 & 72/66 & 72/66 
\\ \cmidrule{2-14}

        & Mixtral 61/72    
            & 77/\underline{75} & 77/\underline{75} & 77/\underline{75} & 77/\underline{75} & 77/\underline{75} & 77/\underline{75} & 77/\underline{75} & 77/75 & 77/\underline{75} & 77/75 & 77/\underline{75} & 77/\underline{75} 
\\

        & Mistral 66/74    
            & 77/\underline{75} & 77/\underline{75} & 77/\underline{75} & 77/\underline{75} & 77/\underline{75} & 77/\underline{75} & 77/\underline{75} & 77/75 & 77/\underline{75} & 77/75 & 77/\underline{75} & 77/\underline{75} 
\\

        & Gemma 72/64       
            & 72/{\bf 76} & 72/{\bf 76} & 72/{\bf 76} & 72/{\bf 76} & 72/{\bf 76} & 72/{\bf 76} & 72/{\bf 76} & 72/\underline{76} & 72/{\bf 76} & 72/\underline{76} & 77/{\bf 76} & 77/{\bf 76} 
\\

        & Gemma-2 78/76     
            & 77/\underline{76} & 77/{\bf 76} & 77/\underline{76} & 77/{\bf 76} & 77/\underline{76} & 77/{\bf 76} & 77/\underline{76} & 77/\underline{76} & 77/\underline{76} & 77/\underline{76} & 77/\underline{76} & 77/{\bf 76} 
\\

        & Llama-3 78/76     
            & 77/\underline{76} & 77/{\bf 76} & 77/\underline{76} & 77/{\bf 76} & 77/\underline{76} & 77/{\bf 76} & 77/\underline{76} & 77/\underline{76} & 77/\underline{76} & 77/\underline{76} & 77/\underline{76} & 77/{\bf 76} 
\\
\end{tabular}
    \caption{Accuracy results of our multi-agent combination framework applied to ArgLLM with Llama-3, RAG-ArgLLM agents with NYTimes (columns), and RAG-ArgLLM agents with Guardian (rows) for Depth 1. We show accuracy results using 0.5/estimated base scores and $\omega_{avg}$ (Avg) and $\omega_{max}$ (Max) base score aggregations. Bold indicates improvement over both Single agents; underline indicates improvement over one and parity with the other.}
    \label{tab:d1_argllm_llama3_nytimes_guardian}
\end{table*}

\begin{table*}[bt]
    \centering
    \begin{tabular}{crcccccccccccc}
        & & \multicolumn{2}{c}{Mixtral} 
          & \multicolumn{2}{c}{Mistral} 
          & \multicolumn{2}{c}{Gemma} 
          & \multicolumn{2}{c}{Gemma-2} 
          & \multicolumn{2}{c}{Llama-3} 
          & \multicolumn{2}{c}{GPT-4o} \\
        \cmidrule(lr){3-4} \cmidrule(lr){5-6} \cmidrule(lr){7-8} 
        \cmidrule(lr){9-10} \cmidrule(lr){11-12} \cmidrule(lr){13-14}
        & & Avg & Max & Avg & Max & Avg & Max & Avg & Max & Avg & Max & Avg & Max \\
        \midrule

        \multirow{7}{*}{GJOpen - 74/69} 
        & Single     
            & 61/71 & 61/71 & 59/68 & 59/68 & 71/66 & 71/66 & 76/75 & 76/75 & 78/44 & 78/44 & 72/66 & 72/66 
\\ \cmidrule{2-14}

        & Mixtral 61/72    
            & {\bf 75}/69 & {\bf 75}/69 & {\bf 75}/69 & {\bf 75}/69 & {\bf 75}/69 & {\bf 75}/69 & 75/69 & 75/69 & 75/69 & 75/69 & {\bf 75}/69 & {\bf 75}/69 
\\

        & Mistral 66/74    
            & \underline{74}/68 & \underline{74}/68 & \underline{74}/68 & \underline{74}/68 & \underline{74}/68 & \underline{74}/68 & 74/68 & 74/68 & 74/68 & 74/68 & \underline{74}/68 & \underline{74}/68 
\\

        & Gemma 72/64       
            & \underline{74}/70 & \underline{74}/70 & \underline{74}/{\bf 70} & \underline{74}/70 & \underline{74}/{\bf 70} & \underline{74}/{\bf 70} & 74/70 & 74/70 & 74/{\bf 70} & 74/70 & \underline{74}/{\bf 70} & \underline{74}/{\bf 70} 
\\

        & Gemma-2 78/76     
            & 74/70 & 74/70 & 74/70 & 74/70 & 74/70 & 74/70 & 74/70 & 74/70 & 74/70 & 74/70 & 74/70 & 74/70 
\\

        & Llama-3 78/76     
            & 74/70 & 74/70 & 74/70 & 74/70 & 74/70 & 74/{\bf 70} & 74/70 & 74/70 & 74/70 & 74/70 & 74/70 & 74/{\bf 70} 
\\
\end{tabular}
    \caption{Accuracy results of our multi-agent combination framework applied to ArgLLM with GPT-4o, RAG-ArgLLM agents with NYTimes (columns), and RAG-ArgLLM agents with Guardian (rows) for Depth 1. We show accuracy results using 0.5/estimated base scores and $\omega_{avg}$ (Avg) and $\omega_{max}$ (Max) base score aggregations. Bold indicates improvement over both Single agents; underline indicates improvement over one and parity with the other.}
    \label{tab:d1_argllm_gpt4o_nytimes_guardian}
\end{table*}

\begin{table*}[bt]
    \centering
    \begin{tabular}{crcccccccccccc}
        & & \multicolumn{2}{c}{Mixtral} 
          & \multicolumn{2}{c}{Mistral} 
          & \multicolumn{2}{c}{Gemma} 
          & \multicolumn{2}{c}{Gemma-2} 
          & \multicolumn{2}{c}{Llama-3} 
          & \multicolumn{2}{c}{GPT-4o} \\
        \cmidrule(lr){3-4} \cmidrule(lr){5-6} \cmidrule(lr){7-8} 
        \cmidrule(lr){9-10} \cmidrule(lr){11-12} \cmidrule(lr){13-14}
        & & Avg & Max & Avg & Max & Avg & Max & Avg & Max & Avg & Max & Avg & Max \\
        \midrule

        \multirow{7}{*}{GJOpen - 56/69} 
        & Single     
            & 49/68 & 49/68 & 51/68 & 51/68 & 64/63 & 64/63 & 77/76 & 77/76 & 74/52 & 74/52 & 72/67 & 72/67 
\\ \cmidrule{2-14}

        & Mixtral 61/72    
            & 28/60 & 42/67 & 28/66 & 43/{\bf 70} & 43/{\bf 70} & 48/{\bf 70} & 53/71 & 48/70 & 26/68 & 33/{\bf 74} & 65/{\bf 73} & 69/{\bf 74} 
\\

        & Mistral 66/74    
            & 30/65 & 44/{\bf 70} & 31/68 & 43/{\bf 71} & 53/{\bf 72} & 48/{\bf 71} & 53/72 & 48/71 & 28/{\bf 71} & 36/{\bf 74} & 69/{\bf 73} & 70/{\bf 74} 
\\

        & Gemma 72/64       
            & 56/{\bf 72} & 47/{\bf 72} & 46/{\bf 73} & 56/{\bf 72} & 55/\underline{69} & 42/{\bf 70} & 55/69 & 42/70 & 49/{\bf 75} & 61/{\bf 74} & 71/{\bf 75} & 70/{\bf 74} 
\\

        & Gemma-2 78/76     
            & 56/72 & 54/69 & 62/74 & 56/72 & 55/69 & 45/69 & 55/69 & 42/65 & 67/{\bf 77} & 61/74 & 71/\underline{75} & 62/73 
\\

        & Llama-3 78/76     
            & 27/68 & 34/{\bf 74} & 30/{\bf 70} & 37/{\bf 74} & 46/{\bf 74} & 41/{\bf 73} & 56/74 & 41/73 & 27/{\bf 73} & 28/{\bf 74} & 64/{\bf 74} & 64/{\bf 74} 
\\
\end{tabular}
    \caption{Accuracy results of our multi-agent combination framework applied to ArgLLM with Mixtral, RAG-ArgLLM agents with NYTimes (columns), and RAG-ArgLLM agents with Guardian (rows) for Depth 2. We show accuracy results using 0.5/estimated base scores and $\omega_{avg}$ (Avg) and $\omega_{max}$ (Max) base score aggregations. Bold indicates improvement over both Single agents; underline indicates improvement over one and parity with the other.}
    \label{tab:d2_argllm_mixtral_nytimes_guardian}
\end{table*}

\begin{table*}[bt]
    \centering
    \begin{tabular}{crcccccccccccc}
        & & \multicolumn{2}{c}{Mixtral} 
          & \multicolumn{2}{c}{Mistral} 
          & \multicolumn{2}{c}{Gemma} 
          & \multicolumn{2}{c}{Gemma-2} 
          & \multicolumn{2}{c}{Llama-3} 
          & \multicolumn{2}{c}{GPT-4o} \\
        \cmidrule(lr){3-4} \cmidrule(lr){5-6} \cmidrule(lr){7-8} 
        \cmidrule(lr){9-10} \cmidrule(lr){11-12} \cmidrule(lr){13-14}
        & & Avg & Max & Avg & Max & Avg & Max & Avg & Max & Avg & Max & Avg & Max \\
        \midrule

        \multirow{7}{*}{GJOpen - 56/67} 
        & Single     
            & 49/68 & 49/68 & 51/68 & 51/68 & 64/63 & 64/63 & 77/76 & 77/76 & 74/52 & 74/52 & 72/67 & 72/67 
\\ \cmidrule{2-14}

        & Mixtral 61/72    
            & 32/59 & 46/67 & 33/62 & 48/{\bf 69} & 53/{\bf 69} & 47/{\bf 71} & 53/69 & 50/68 & 33/66 & 39/{\bf 73} & 64/{\bf 74} & 68/{\bf 75} 
\\

        & Mistral 66/74    
            & 34/60 & 47/{\bf 69} & 34/63 & 46/{\bf 69} & 44/{\bf 69} & 46/{\bf 71} & 52/69 & 49/68 & 34/67 & 40/{\bf 73} & 67/{\bf 74} & 69/{\bf 75} 
\\

        & Gemma 72/64       
            & 45/{\bf 69} & 53/\underline{68} & 60/{\bf 72} & 49/{\bf 71} & 56/{\bf 70} & 49/{\bf 68} & 56/70 & 45/62 & 51/{\bf 73} & 46/{\bf 74} & 70/{\bf 75} & 65/{\bf 74} 
\\

        & Gemma-2 78/76     
            & 56/70 & 53/68 & 60/72 & 56/71 & 61/71 & 49/68 & 56/67 & 45/62 & 68/{\bf 76} & 59/\underline{75} & 71/{\bf 76} & 59/73 
\\

        & Llama-3 78/76     
            & 34/65 & 41/{\bf 73} & 35/\underline{68} & 42/{\bf 73} & 47/{\bf 72} & 42/{\bf 73} & 56/73 & 44/71 & 35/{\bf 71} & 34/{\bf 73} & 69/{\bf 75} & 67/{\bf 76} 
\\
\end{tabular}
    \caption{Accuracy results of our multi-agent combination framework applied to ArgLLM with Mistral, RAG-ArgLLM agents with NYTimes (columns), and RAG-ArgLLM agents with Guardian (rows) for Depth 2. We show accuracy results using 0.5/estimated base scores and $\omega_{avg}$ (Avg) and $\omega_{max}$ (Max) base score aggregations. Bold indicates improvement over both Single agents; underline indicates improvement over one and parity with the other.}
    \label{tab:d2_argllm_mistral_nytimes_guardian}
\end{table*}

\begin{table*}[bt]
    \centering
    \begin{tabular}{crcccccccccccc}
        & & \multicolumn{2}{c}{Mixtral} 
          & \multicolumn{2}{c}{Mistral} 
          & \multicolumn{2}{c}{Gemma} 
          & \multicolumn{2}{c}{Gemma-2} 
          & \multicolumn{2}{c}{Llama-3} 
          & \multicolumn{2}{c}{GPT-4o} \\
        \cmidrule(lr){3-4} \cmidrule(lr){5-6} \cmidrule(lr){7-8} 
        \cmidrule(lr){9-10} \cmidrule(lr){11-12} \cmidrule(lr){13-14}
        & & Avg & Max & Avg & Max & Avg & Max & Avg & Max & Avg & Max & Avg & Max \\
        \midrule

        \multirow{7}{*}{GJOpen - 70/56} 
        & Single     
            & 49/68 & 49/68 & 51/68 & 51/68 & 64/63 & 64/63 & 77/76 & 77/76 & 74/52 & 74/52 & 72/67 & 72/67 
\\ \cmidrule{2-14}

        & Mixtral 61/72    
            & 45/\underline{68} & 47/67 & 47/{\bf 71} & 50/{\bf 71} & 54/52 & 50/{\bf 74} & 63/75 & 49/52 & 50/50 & 41/{\bf 77} & 56/50 & 49/{\bf 72} 
\\

        & Mistral 66/74    
            & 45/{\bf 69} & 47/47 & 45/48 & 46/47 & 62/54 & 47/49 & 62/54 & 54/70 & 53/{\bf 76} & 40/{\bf 76} & 57/51 & 53/48 
\\

        & Gemma 72/64       
            & 51/51 & 49/65 & 57/{\bf 72} & 47/{\bf 72} & {\bf 73}/{\bf 77} & 54/53 & 73/{\bf 77} & 68/74 & \underline{74}/{\bf 78} & 44/{\bf 72} & \underline{72}/56 & 45/48 
\\

        & Gemma-2 78/76     
            & 60/55 & 49/65 & 57/72 & 49/65 & {\bf 76}/{\bf 76} & 54/53 & 76/\underline{76} & 68/74 & 74/{\bf 78} & 44/72 & 72/56 & 45/48 
\\

        & Llama-3 78/76     
            & 49/50 & 43/{\bf 77} & 54/{\bf 76} & 43/{\bf 77} & 70/59 & 44/48 & 70/59 & 48/75 & 66/{\bf 77} & 31/48 & 73/{\bf 77} & 45/48 
\\
\end{tabular}
    \caption{Accuracy results of our multi-agent combination framework applied to ArgLLM with Gemma, RAG-ArgLLM agents with NYTimes (columns), and RAG-ArgLLM agents with Guardian (rows) for Depth 2. We show accuracy results using 0.5/estimated base scores and $\omega_{avg}$ (Avg) and $\omega_{max}$ (Max) base score aggregations. Bold indicates improvement over both Single agents; underline indicates improvement over one and parity with the other.}
    \label{tab:d2_argllm_gemma_nytimes_guardian}
\end{table*}

\begin{table*}[bt]
    \centering
    \begin{tabular}{crcccccccccccc}
        & & \multicolumn{2}{c}{Mixtral} 
          & \multicolumn{2}{c}{Mistral} 
          & \multicolumn{2}{c}{Gemma} 
          & \multicolumn{2}{c}{Gemma-2} 
          & \multicolumn{2}{c}{Llama-3} 
          & \multicolumn{2}{c}{GPT-4o} \\
        \cmidrule(lr){3-4} \cmidrule(lr){5-6} \cmidrule(lr){7-8} 
        \cmidrule(lr){9-10} \cmidrule(lr){11-12} \cmidrule(lr){13-14}
        & & Avg & Max & Avg & Max & Avg & Max & Avg & Max & Avg & Max & Avg & Max \\
        \midrule

        \multirow{7}{*}{GJOpen - 79/77} 
        & Single     
            & 49/68 & 49/68 & 51/68 & 51/68 & 64/63 & 64/63 & 77/76 & 77/76 & 74/52 & 74/52 & 72/67 & 72/67 
\\ \cmidrule{2-14}

        & Mixtral 61/72    
            & 45/68 & 47/67 & 47/71 & 50/71 & 63/75 & 50/74 & 63/75 & 53/71 & 53/76 & 41/\underline{77} & 55/73 & 49/72 
\\

        & Mistral 66/74    
            & 45/69 & 49/71 & 47/72 & 49/73 & 57/75 & 54/70 & 63/74 & 54/70 & 53/76 & 40/76 & 56/75 & 49/75 
\\

        & Gemma 72/64       
            & 54/73 & 49/65 & 57/72 & 47/72 & 73/\underline{77} & 68/74 & 73/\underline{77} & 68/74 & 74/{\bf 78} & 44/72 & 72/\underline{77} & 46/73 
\\

        & Gemma-2 78/76     
            & 56/71 & 49/65 & 57/72 & 49/65 & 76/76 & 68/74 & 76/76 & 68/74 & 74/{\bf 78} & 44/72 & 72/\underline{77} & 46/73 
\\

        & Llama-3 78/76     
            & 52/75 & 43/\underline{77} & 54/76 & 43/\underline{77} & 68/\underline{77} & 48/75 & 73/\underline{77} & 48/75 & 66/\underline{77} & 32/\underline{77} & 73/\underline{77} & 45/76 
\\
\end{tabular}
    \caption{Accuracy results of our multi-agent combination framework applied to ArgLLM with Gemma-2, RAG-ArgLLM agents with NYTimes (columns), and RAG-ArgLLM agents with Guardian (rows) for Depth 2. We show accuracy results using 0.5/estimated base scores and $\omega_{avg}$ (Avg) and $\omega_{max}$ (Max) base score aggregations. Bold indicates improvement over both Single agents; underline indicates improvement over one and parity with the other.}
    \label{tab:d2_argllm_gemma2_nytimes_guardian}
\end{table*}

\begin{table*}[bt]
    \centering
    \begin{tabular}{crcccccccccccc}
        & & \multicolumn{2}{c}{Mixtral} 
          & \multicolumn{2}{c}{Mistral} 
          & \multicolumn{2}{c}{Gemma} 
          & \multicolumn{2}{c}{Gemma-2} 
          & \multicolumn{2}{c}{Llama-3} 
          & \multicolumn{2}{c}{GPT-4o} \\
        \cmidrule(lr){3-4} \cmidrule(lr){5-6} \cmidrule(lr){7-8} 
        \cmidrule(lr){9-10} \cmidrule(lr){11-12} \cmidrule(lr){13-14}
        & & Avg & Max & Avg & Max & Avg & Max & Avg & Max & Avg & Max & Avg & Max \\
        \midrule

        \multirow{7}{*}{GJOpen - 63/75} 
        & Single     
            & 49/68 & 49/68 & 51/68 & 51/68 & 64/63 & 64/63 & 77/76 & 77/76 & 74/52 & 74/52 & 72/67 & 72/67 
\\ \cmidrule{2-14}

        & Mixtral 61/72    
            & 26/65 & 36/74 & 28/70 & 39/\underline{75} & 44/74 & 43/\underline{75} & 52/74 & 42/75 & 26/74 & 28/\underline{75} & 53/74 & 62/\underline{75} 
\\

        & Mistral 66/74    
            & 30/69 & 39/\underline{75} & 31/73 & 39/\underline{75} & 44/74 & 42/\underline{75} & 52/75 & 41/74 & 28/\underline{75} & 32/\underline{75} & 61/\underline{75} & 64/\underline{75} 
\\

        & Gemma 72/64       
            & 60/\underline{75} & 54/71 & {\bf 65}/{\bf 76} & 55/\underline{75} & 56/{\bf 76} & 47/72 & 56/\underline{76} & 39/74 & 70/{\bf 78} & 65/\underline{75} & 67/{\bf 76} & 62/\underline{75} 
\\

        & Gemma-2 78/76     
            & 60/\underline{75} & 54/71 & 65/{\bf 76} & 55/\underline{75} & 56/{\bf 76} & 47/72 & 56/\underline{76} & 40/71 & 70/{\bf 78} & 65/\underline{75} & 72/{\bf 77} & 53/\underline{75} 
\\

        & Llama-3 78/76     
            & 27/73 & 30/\underline{75} & 30/\underline{75} & 34/\underline{75} & 47/\underline{75} & 33/\underline{75} & 57/75 & 28/74 & 26/\underline{75} & 25/\underline{75} & 65/\underline{75} & 66/\underline{75} 
\\
\end{tabular}
    \caption{Accuracy results of our multi-agent combination framework applied to ArgLLM with Llama-3, RAG-ArgLLM agents with NYTimes (columns), and RAG-ArgLLM agents with Guardian (rows) for Depth 2. We show accuracy results using 0.5/estimated base scores and $\omega_{avg}$ (Avg) and $\omega_{max}$ (Max) base score aggregations. Bold indicates improvement over both Single agents; underline indicates improvement over one and parity with the other.}
    \label{tab:d2_argllm_llama3_nytimes_guardian}
\end{table*}

\begin{table*}[bt]
    \centering
    \begin{tabular}{crcccccccccccc}
        & & \multicolumn{2}{c}{Mixtral} 
          & \multicolumn{2}{c}{Mistral} 
          & \multicolumn{2}{c}{Gemma} 
          & \multicolumn{2}{c}{Gemma-2} 
          & \multicolumn{2}{c}{Llama-3} 
          & \multicolumn{2}{c}{GPT-4o} \\
        \cmidrule(lr){3-4} \cmidrule(lr){5-6} \cmidrule(lr){7-8} 
        \cmidrule(lr){9-10} \cmidrule(lr){11-12} \cmidrule(lr){13-14}
        & & Avg & Max & Avg & Max & Avg & Max & Avg & Max & Avg & Max & Avg & Max \\
        \midrule

        \multirow{7}{*}{GJOpen - 69/68} 
        & Single     
            & 49/68 & 49/68 & 51/68 & 51/68 & 64/63 & 64/63 & 77/76 & 77/76 & 74/52 & 74/52 & 72/67 & 72/67 
\\ \cmidrule{2-14}

        & Mixtral 61/72    
            & 26/55 & 43/66 & 28/57 & 43/66 & 61/\underline{68} & 42/66 & 61/68 & 44/66 & 26/62 & 32/65 & 71/\underline{68} & \underline{72}/\underline{68} 
\\

        & Mistral 66/74    
            & 28/57 & 42/65 & 31/61 & 41/66 & 48/66 & 42/65 & 61/68 & 42/65 & 28/63 & 33/65 & \underline{72}/\underline{68} & \underline{72}/\underline{68} 
\\

        & Gemma 72/64       
            & 61/{\bf 69} & 47/64 & 66/{\bf 71} & 54/66 & 62/\underline{68} & 33/65 & 62/68 & 30/62 & 50/\underline{68} & 61/67 & {\bf 75}/{\bf 74} & 71/\underline{68} 
\\

        & Gemma-2 78/76     
            & 61/69 & 53/62 & 66/71 & 54/66 & 59/69 & 39/63 & 59/69 & 30/62 & 70/72 & 61/67 & \underline{75}/74 & 65/68 
\\

        & Llama-3 78/76     
            & 27/61 & 33/65 & 28/63 & 35/66 & 66/\underline{68} & 38/66 & 66/68 & 40/65 & 27/65 & 26/65 & 72/{\bf 69} & 70/{\bf 69} 
\\
\end{tabular}
    \caption{Accuracy results of our multi-agent combination framework applied to ArgLLM with GPT-4o, RAG-ArgLLM agents with NYTimes (columns), and RAG-ArgLLM agents with Guardian (rows) for Depth 2. We show accuracy results using 0.5/estimated base scores and $\omega_{avg}$ (Avg) and $\omega_{max}$ (Max) base score aggregations. Bold indicates improvement over both Single agents; underline indicates improvement over one and parity with the other.}
    \label{tab:d2_argllm_gpt4o_nytimes_guardian}
\end{table*}

{}
}

\end{document}

%% file: macros.tex
\usepackage{savesym}
\savesymbol{Bbbk}
\savesymbol{openbox}

\restoresymbol{NEW}{Bbbk}
\restoresymbol{NEW}{openbox}
\usepackage{tcolorbox}

\usepackage{booktabs}
\usepackage{todonotes}
\usepackage{amsthm}
\usepackage{algorithm}
\usepackage[noend]{algorithmic}
\usepackage{multirow}
\usepackage{multicol}
\usepackage{thmtools}

\newtheorem{definition}{Definition}
\newtheorem{example}{Example}
\declaretheorem[name=Proposition]{proposition}

\newcommand{\note}[1]{\textcolor{magenta}{$*$ #1}}
\newcommand{\delete}[1]{\textcolor{olive}{$\times$ #1}}

\newcommand{\Pro}{\mathrm{Pro}}
\newcommand{\Con}{\mathrm{Con}}
\newcommand{\Path}{\mathrm{path}}

\newboolean{arxivversion}
\setboolean{arxivversion}{true} 
